\documentclass[letterpaper]{article} % DO NOT CHANGE THIS
\usepackage[preprint]{aaai2027}  % DO NOT CHANGE THIS
\usepackage[hyphens]{url}  % DO NOT CHANGE THIS
\usepackage{graphicx} % DO NOT CHANGE THIS
\usepackage{natbib}  % DO NOT CHANGE THIS AND DO NOT ADD ANY OPTIONS TO IT
\usepackage{caption} % DO NOT CHANGE THIS AND DO NOT ADD ANY OPTIONS TO IT
\usepackage{algorithm}
\usepackage{algorithmic}

\usepackage{newfloat}
\usepackage{listings}
\DeclareCaptionStyle{ruled}{labelfont=normalfont,labelsep=colon,strut=off} % DO NOT CHANGE THIS
\floatstyle{ruled}
\newfloat{listing}{tb}{lst}{}
\floatname{listing}{Listing}

\usepackage{booktabs}

\usepackage{amsmath}
\usepackage{amssymb}
\usepackage{subcaption}
\usepackage{array}
\usepackage{tabularx}

\usepackage{xcolor}
\usepackage{multirow}
\usepackage{colortbl}
\usepackage{subcaption}
\title{A Frozen Pixel-Space Diffusion Model Can Guide Itself with Its Own Samples}

\author{
Zixuan Fu,
Chong Wang,
Lanqing Guo,
Kailai Zhou,
Jiahao Nie,
Bihan Wen\corresponding
}

\affiliations{
Nanyang Technological University
}

\begin{document}

\maketitle

\begin{abstract}
Pixel-space diffusion models aim to learn an end-to-end generator directly over raw pixels.
This is challenging because a single model must capture both global structure and local texture in the same high-dimensional space.
While recent work improves pixel diffusion through alternative prediction targets, training objectives, and architectures, these advances typically require training a new model from scratch.
We show there is a cheaper, complementary strategy: \textbf{a frozen, pretrained pixel diffusion model can guide itself}. 
Our key observation is that intermediate layers of a pretrained pixel diffusion transformer can be decoded into coarse predictions that capture the main low-frequency structure, while the final layers progressively refine local, high-frequency details.
We therefore attach a lightweight prediction head to an intermediate layer, keep the backbone frozen, and use the discrepancy between the intermediate and final predictions as a self-guidance direction during sampling.
To train this head, we further find that real images are not necessary.
Instead, model-generated samples suffice and even outperform real images for training the head, especially in enhancing the high-frequency components that pixel diffusion tends to underfit.
Across multiple pixel diffusion models on ImageNet, our \textbf{Synthetic Self-Guidance (SSG)} consistently improves generation while adapter training requires less than 1$\%$ of full-model training compute: it reduces FID by over 50$\%$ across the evaluated JiT variants without classifier-free guidance (CFG) and further improves strong baselines with CFG, e.g., JiT-H/16 from 1.86 to 1.67 and PixelREPA-H/16 from 1.81 to 1.59.
Our code is available at \url{https://github.com/zfu006/SSG}.
\end{abstract}

\section{Introduction}
Pixel-space diffusion models generate images directly in raw pixel space \cite{li2026back,yu2026pixeldit,dhariwal2021diffusion}.
Unlike latent diffusion models (LDMs) \cite{rombach2022high}, they avoid the two-stage pipeline of training and using a VAE to compress images into latent representations.
However, this simplicity comes with a harder modeling problem: a single end-to-end model must capture both global semantics and fine-grained local details in the same high-dimensional space.
Recent progress has improved pixel diffusion by shifting the prediction target from noise or velocity to clean images, i.e., $x$-prediction \cite{li2026back}, and by adding perceptual supervision, representation alignment \cite{ma2026pixelgen,shin2026representation,yu2024representation,lei2025there}, or hierarchical architectures that decouple semantic structure and high-frequency components \cite{yu2026pixeldit,ma2026deco,guo2026pixelu}.
Despite these advances, most existing methods require designing and training a new model from scratch, which is computationally expensive and does not fully exploit the prior knowledge of a pretrained pixel diffusion model.

 \begin{figure}[t]
  \centering
  \setlength{\tabcolsep}{0pt}

  \begin{tabularx}{\columnwidth}{
  @{}
  *{3}{>{\centering\arraybackslash}X}
  @{}
  }
  {\small No Guidance} &
  {\small CFG} &
  {\small CFG + SSG}
  \\

  \multicolumn{3}{@{}c@{}}{%
    \includegraphics[
      width=\columnwidth
    ]{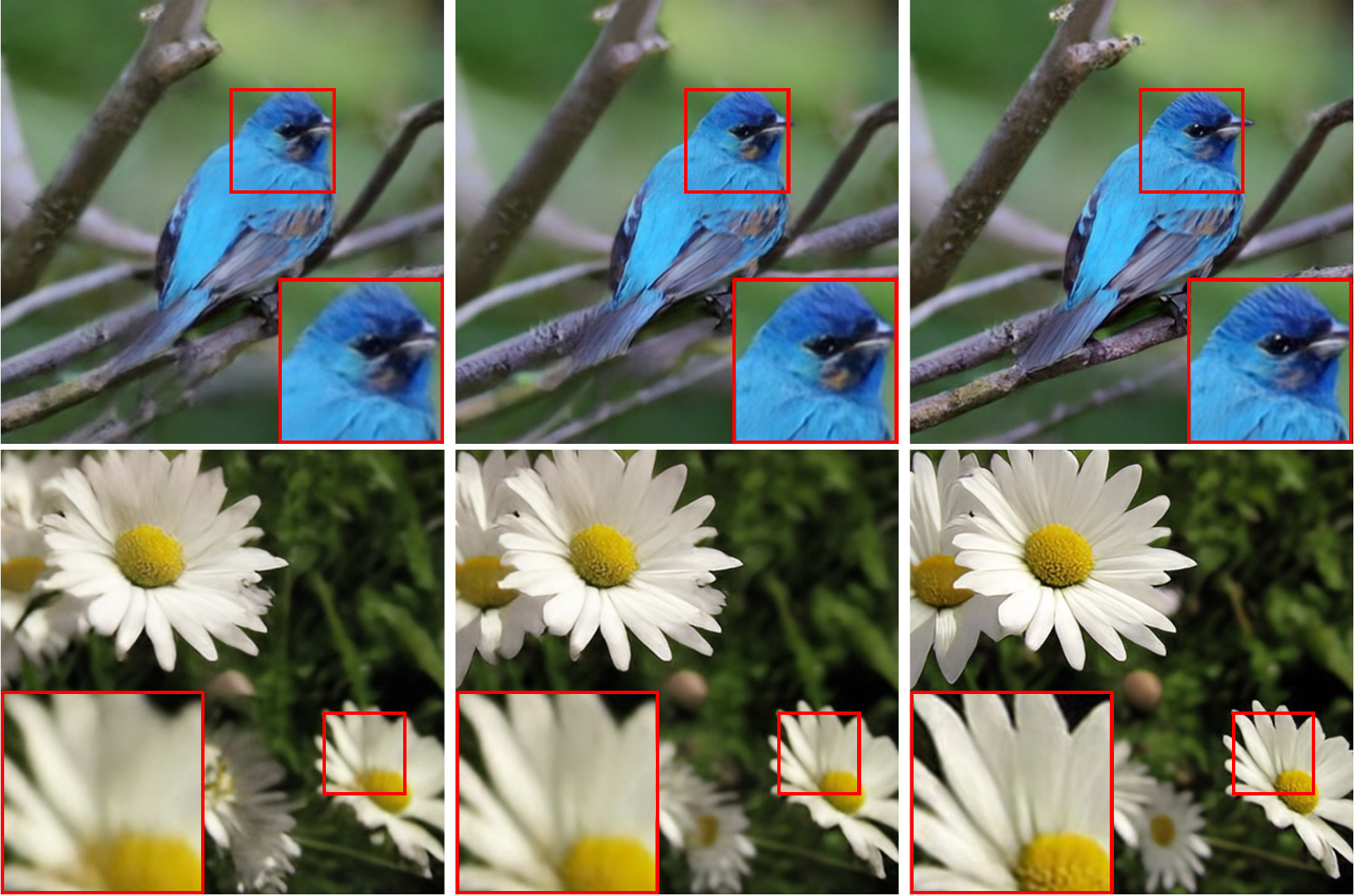}
  }
  \end{tabularx}

  \caption{Qualitative comparison of 
  guidance strategies using JiT-H/32. SSG refines local textures and fine-grained details.}
  \label{fig_guidance_comparison}
  \end{figure}

This raises a natural question: \textit{Can a pretrained pixel diffusion model use its own learned representations to improve generation without retraining the backbone?}
Our key observation is that intermediate representations of a plain pixel diffusion transformer, such as JiT \cite{li2026back}, can be decoded into coarse predictions that capture the main low-frequency structure, while the final layers progressively refine local, high-frequency details.
We then turn this internal coarse-to-fine progression into a guidance signal: we attach a lightweight adapter to an intermediate layer, keep the backbone frozen, and extrapolate the final prediction away from the coarse intermediate prediction during sampling, pushing further along the model's own refinement direction and thereby enhancing high-frequency details that pixel diffusion models struggle to fit.

The remaining design choice is how to train the intermediate adapter.
Intuitively, the adapter only needs to provide a coarse, low-frequency prediction as a weak reference for self-guidance, rather than to synthesize the full frequency components of real images.
We therefore propose to train the adapter on model-generated samples rather than real images, so that the coarse prediction focuses on low-frequency structure and provides effective guidance that helps alleviate blurry textures and missing high-frequency detail in the final generated images (Fig.~\ref{fig_guidance_comparison}).
Surprisingly, training the adapter on synthetic samples leads to better generation performance than training it on real images.
We refer to the complete framework as \textbf{Synthetic Self-Guidance} (SSG), a plug-in strategy that improves pretrained pixel diffusion models without requiring real data for adapter training, as illustrated in Fig.~\ref{fig_ssg_method}.
We evaluate SSG across multiple pixel diffusion models (JiT \cite{li2026back}, PixelREPA \cite{shin2026representation}, and DeCo \cite{ma2026deco}) on ImageNet \cite{russakovsky2015imagenet} at 256 and 512 resolutions. 
Although training the adapter costs less than 1$\%$ of the full model's training compute, SSG reduces FID by over 50$\%$ across the evaluated JiT variants without classifier-free guidance (CFG), and further improves strong baselines with CFG, e.g., JiT-H/16 from 1.86 to 1.67 and PixelREPA-H/16 from 1.81 to 1.59.
Our contributions are threefold:
\begin{itemize}
\item We show that intermediate representations of a pretrained pixel diffusion transformer can be decoded into coarse predictions, while final layers progressively refine high-frequency details. We turn their discrepancy into guidance during sampling without retraining the backbone.
\item We find that the adapter can be trained without any real data: the model's own generated samples are even more effective than real images for training the adapter.
\item We propose \textbf{Synthetic Self-Guidance} (SSG), a plug-in method that trains only the adapter using less than 1$\%$ of the full model's training compute, yet improves several pixel diffusion models to competitive FID on ImageNet (e.g., 1.59 on PixelREPA-H/16).
\end{itemize}

% \begin{figure*}
%     \includegraphics[width=\linewidth]{Figures/tesr.png}
% \end{figure*}
\section{Related Work}
\subsection{Latent and Pixel-Space Diffusion Models}
Latent diffusion models (LDMs) typically use a separate VAE to compress images into a lower-dimensional latent space, reducing computational cost by training the generative diffusion model in this latent space \cite{rombach2022high}.
DiT replaces the U-Net \cite{ronneberger2015u} with a transformer architecture, while SiT further adopts velocity prediction and flow matching
\cite{peebles2023scalable, ma2024sit, ho2020denoising, liu2022flow}.
Recent methods such as REPA \cite{yu2024representation}, VA-VAE \cite{yao2025reconstruction}, and RAE \cite{zheng2025diffusion} adopt pretrained visual representations from foundation models to improve generation in latent space.
While LDMs are effective, they retain a two-stage pipeline, and the VAE bottleneck can limit reconstruction fidelity and consequently generation quality.

Pixel-space diffusion models instead perform denoising directly over raw pixels.
The original DDPMs \cite{ho2020denoising} apply U-Nets to predict noise, but their computational cost increases substantially with image resolution.
JiT \cite{li2026back} adopts a plain Vision Transformer (ViT) \cite{dosovitskiy2020image} with large patches and directly predicts clean images, reducing the difficulty of modeling high-dimensional raw-pixel patches.
DeCo \cite{ma2026deco}, DiP \cite{chen2026dip}, and PixelDiT \cite{yu2026pixeldit} introduce hierarchical architectures that better decouple semantic structure and high-frequency details for generation.
Representation learning and alignment \cite{yu2024representation} have also been explored in pixel diffusion models.
PixelREPA \cite{shin2026representation} shows that directly distilling representations from vision foundation models can conflict with pixel-space denoising and addresses this issue with a masked transformer adapter for representation alignment.
EPG \cite{lei2025there} instead adopts an encoder pretrained through self-supervised representation learning and jointly trains it with a decoder for denoising.
PixelGen \cite{ma2026pixelgen} introduces perceptual supervision \cite{johnson2016perceptual} to improve generation quality and convergence.
Other approaches combine latent and pixel diffusion, as in Latent Forcing \cite{baade2026latent}, or introduce a U-shaped architecture, as in PixelU \cite{guo2026pixelu}.
These methods improve pixel diffusion through new objectives, supervision, or architectures, whereas SSG improves an existing pretrained model while keeping its backbone frozen.

\subsection{Guidance for Diffusion Models}
Guidance during sampling is important for improving the quality of diffusion generation. 
Classifier-free guidance (CFG) \cite{ho2022classifier} combines conditional and unconditional predictions by treating the unconditional prediction as a negative reference and extrapolating toward the conditional prediction.
Autoguidance (AG) \cite{karras2024guiding} further replaces unconditional branches in the CFG with a weak model, using its output to guide the stronger model.
Internal Guidance (IG) \cite{zhou2026guiding} removes the need for a separate weak model by jointly training the diffusion model with an additional output head at an intermediate layer, whose output serves as a weak prediction.
Recently, RAEv2 \cite{singh2026improved} proposes to align its internal representation with vision foundation models and applies the aligned representations to guide the model in the representation space.
SSG is most closely related to IG because both obtain weak and strong predictions from the same backbone.
The key difference is that IG obtains this intermediate prediction by jointly training the head together with the backbone, while SSG keeps a pretrained backbone frozen and trains only a lightweight adapter on the model's own synthetic samples, making it a plug-in that requires neither backbone retraining nor real data.

\begin{figure*}[t!]
\centering
\includegraphics[width=\linewidth]{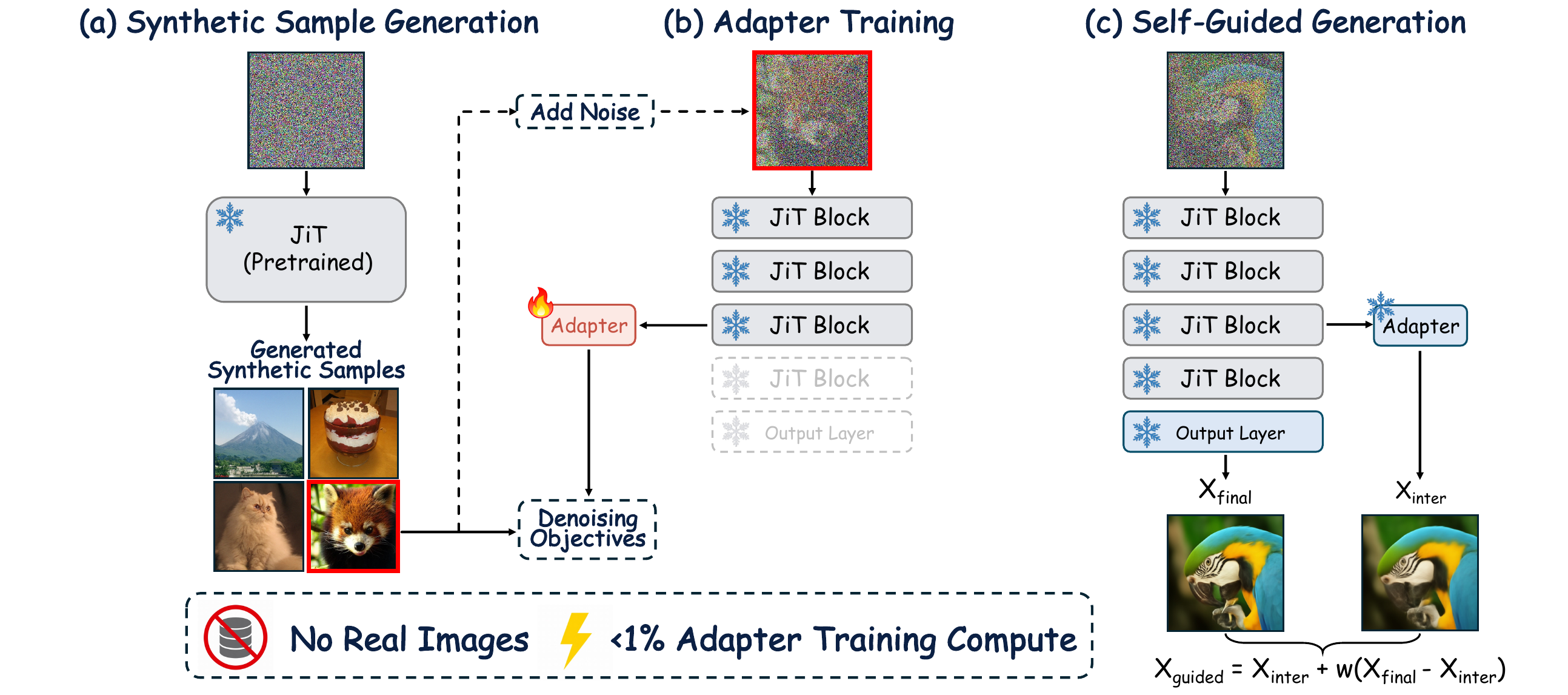}
\caption{Overview of Synthetic Self-Guidance (SSG). 
(a) We sample synthetic images from a pretrained pixel diffusion transformer. 
(b) A lightweight adapter, attached to an intermediate layer, is trained on the synthetic samples while the backbone remains frozen.
(c) During sampling, the intermediate prediction is used to guide the final output.}
\label{fig_ssg_method}
\end{figure*}
\subsection{Self-Improving Diffusion with Synthetic Data}
Recent studies explore using synthetic samples generated by the model itself as training or negative signals to improve pretrained generative models.
SIMS \cite{alemohammad2024self} initializes an auxiliary score model from the base model, trains it on self-synthesized samples, and uses its score as negative guidance during generation to avoid reproducing degraded patterns from these samples.
Neon \cite{alemohammad2025neon} briefly fine-tunes the model on its own samples and then negatively extrapolates in weight space.
SSG also uses synthetic samples but is motivated differently: our starting point is the internal coarse-to-fine structure of a pretrained pixel diffusion model, where synthetic samples serve only as training data for its coarse intermediate prediction. 
SSG therefore keeps the pretrained model frozen, trains only a lightweight intermediate adapter, and uses the discrepancy between the intermediate and final predictions as guidance, requiring neither a separate weak model nor backbone fine-tuning.

\section{Method}
In this section, we present \textbf{Synthetic Self-Guidance} (SSG), a plug-in method that improves a pretrained pixel diffusion model using its synthetic samples.
We first review the formulation and training objective of pixel diffusion models.
We then introduce how an internal representation can be decoded into a coarse prediction with a lightweight adapter and used to guide the final prediction during sampling. 
Finally, we describe how the inserted intermediate adapter is trained entirely with samples generated by the pretrained model and show that this strategy yields better performance than training it on real images.

\subsection{Preliminaries}
\paragraph{Diffusion Models and Flow Matching.} 
Diffusion models learn to generate images through iterative denoising. 
Previous methods such as DDPMs \cite{ho2020denoising} train neural networks to predict noise, while recent flow matching methods \cite{liu2022flow, lipman2022flow} formulate generation by learning a time-dependent velocity field:
\begin{equation}
\label{eq_fm}
  \mathcal{L}_{\mathrm{FM}}
  =
  \mathbb{E}_{\mathbf{x},\boldsymbol{\epsilon},t}
  \left[
  \left\|
  \mathbf{v}_{\theta}(\mathbf{x}_t,t)
  -
  \mathbf{v}_t
  \right\|_2^2
  \right],
\end{equation}
where $\mathbf{v}_{\theta}$ denotes the velocity field predicted by the diffusion model and $\mathbf{v}_t$ is the target velocity at timestep $t$.
Here, $\mathbf{x}_t$ is interpolated between a clean image $\mathbf{x}$ and Gaussian noise $\boldsymbol{\epsilon}$:
\begin{equation}
\label{eq_interpolated}
  \mathbf{x}_t
  =
  t\mathbf{x}
  +
  (1-t)\boldsymbol{\epsilon},
  \quad
  \mathbf{v}_t
  =
  \frac{\mathrm{d}\mathbf{x}_t}{\mathrm{d}t}
  =
  \mathbf{x}-\boldsymbol{\epsilon}.
\end{equation}
JiT adopts a plain transformer architecture to model the image distribution directly in pixel space.
To alleviate the difficulty of predicting noise over large patches in high-dimensional pixel space, JiT directly predicts the clean image:
\begin{equation}
  \hat{\mathbf{x}}
  =
  f_{\theta}(\mathbf{x}_t,t).
\end{equation}
The clean prediction $\hat{\mathbf{x}}$ is then converted into velocity:
\begin{equation}
    \label{eq_x_to_v}
    \mathbf{v}_{\theta}(\mathbf{x}_t,t)
    =
    \frac{
      \hat{\mathbf{x}}-\mathbf{x}_t
    }{
      1-t
    }.
\end{equation}
The diffusion model is trained using the objective in Eq.~\ref{eq_fm}.

\begin{figure}[t]
  \centering
  \begin{subfigure}[t]{0.49\linewidth}
      \centering
      \includegraphics[width=\linewidth]{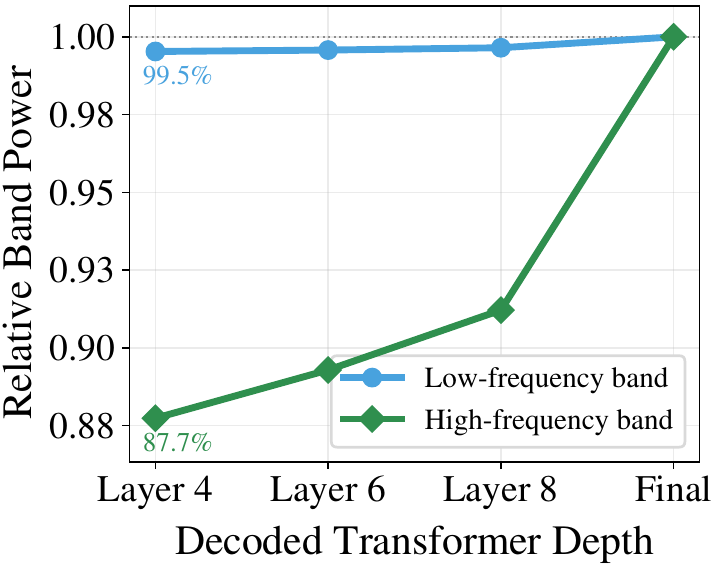}
      \caption{Layer-wise band power}
      \label{fig:ssg_layerwise_frequency}
  \end{subfigure}
  \hfill
  \begin{subfigure}[t]{0.49\linewidth}
      \centering
      \includegraphics[width=\linewidth]{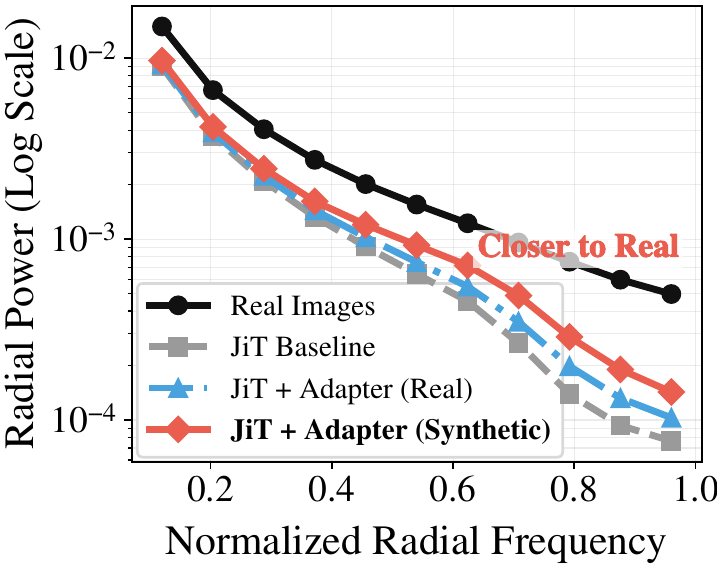}
      \caption{Radial power comparison}
      \label{fig:ssg_radial_spectrum}
  \end{subfigure}

  \caption{{Frequency-domain analysis.}
  (a) Low-frequency and high-frequency band power of intermediate predictions across network layers, normalized by that of the final prediction. 
  (b) Radial power spectra of real images, samples from the JiT-B/16 baseline, and samples generated with self-guidance using adapters trained on real or synthetic images.}
  \label{fig_frequency_analysis}
\end{figure}

\paragraph{Autoguidance and Internal Guidance.} 
Autoguidance (AG) \cite{karras2024guiding} and Internal Guidance (IG) \cite{zhou2026guiding} use a weak
prediction together with the strong prediction to guide generation:
\begin{equation}
\label{eq_weak_guidance}
  \mathbf{v}_{\mathrm{guided}}
  =
  \mathbf{v}_{\mathrm{w}}
  +
  w
  \left(
  \mathbf{v}_{\mathrm{s}}
  -
  \mathbf{v}_{\mathrm{w}}
  \right),
\end{equation}
where $\mathbf{v}_{\mathrm{w}}$ and $\mathbf{v}_{\mathrm{s}}$ denote the weak and strong predictions, respectively, and $w$ is the guidance scale.
AG obtains the weak prediction from a separate weak model, while IG uses an internal output from the diffusion model itself.

\subsection{Synthetic Self-Guidance for Pixel Diffusion}
\paragraph{Observation and Motivation.} 
Our goal is to improve generation using the pretrained pixel diffusion transformer itself, without training a new model from scratch.
We first examine what an intermediate layer already encodes about the clean image by attaching a lightweight trainable adapter to the corresponding layer of a pretrained JiT-B/16 model at 256$\times$256 resolution while keeping the backbone frozen.
The adapter decodes the intermediate representation back into pixel space, producing an intermediate clean prediction.
For this analysis, we independently train an adapter at each of Layers 4, 6, and 8 on the ImageNet training set using the same flow-matching objective as JiT, while updating only the adapter parameters.
We sample 512 images from the ImageNet validation set, construct noisy inputs at $t=0.5$, and obtain the corresponding intermediate clean predictions together with the final prediction.
For each prediction, we apply a two-dimensional Fourier transform to each RGB channel and compute its power spectrum by summing the squared magnitudes across channels.
We then aggregate the power over radial frequency bands, exclude the DC component, and divide the spectrum into a low-frequency band $(0,0.5)$ and a high-frequency band $[0.5,1]$, with the radial frequency normalized such that $1$ corresponds to the Nyquist frequency.
The power within each band is normalized by the corresponding band power of the final prediction.

\begin{figure}[t]
\centering
\setlength{\tabcolsep}{0.5pt}

\begin{tabular}{
    @{}
    >{\centering\arraybackslash}m{0.035\columnwidth}
    *{4}{>{\centering\arraybackslash}m{0.235\columnwidth}}
    @{}
}
    &
    {\small Noisy input} &
    {\small Intermediate} &
    {\small Final} &
    {\small Residual}
    \\[0mm]

    \parbox[c][0.235\columnwidth][c]{\linewidth}{
      \centering
      \rotatebox[origin=c]{90}{\small $t=0.3$}
  } &
    \raisebox{-0.5\height}{\includegraphics[width=\linewidth]{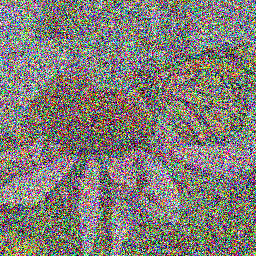}} &
    \raisebox{-0.5\height}{\includegraphics[width=\linewidth]{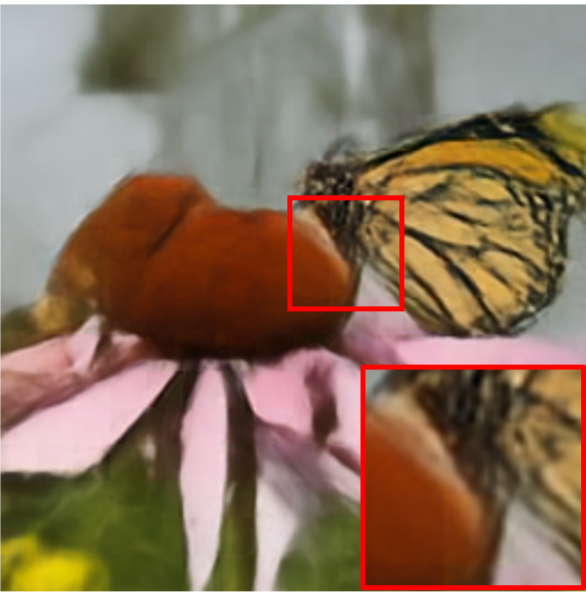}} &
    \raisebox{-0.5\height}{\includegraphics[width=\linewidth]{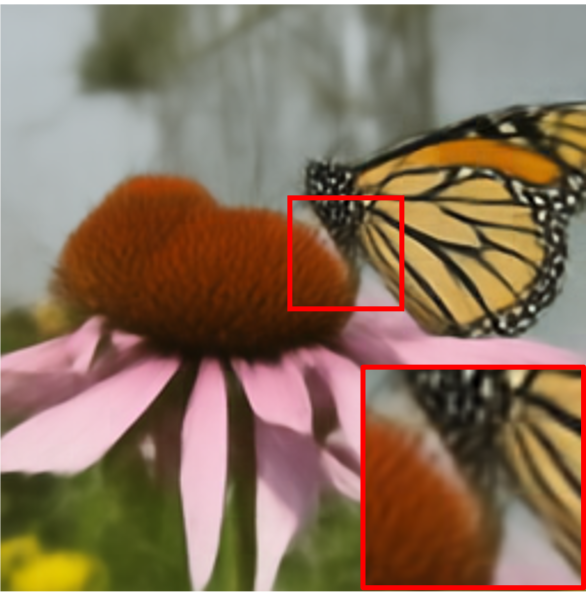}} &
    \raisebox{-0.5\height}{\includegraphics[width=\linewidth]{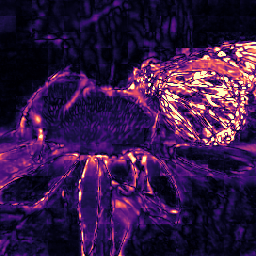}}
    \\[2.2mm]

    \parbox[c][0.235\columnwidth][c]{\linewidth}{
      \centering
      \rotatebox[origin=c]{90}{\small $t=0.6$}
  } &
    \raisebox{-0.5\height}{\includegraphics[width=\linewidth]{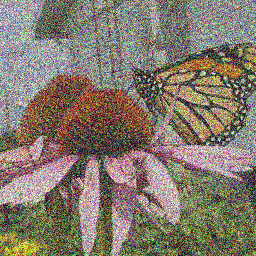}} &
    \raisebox{-0.5\height}{\includegraphics[width=\linewidth]{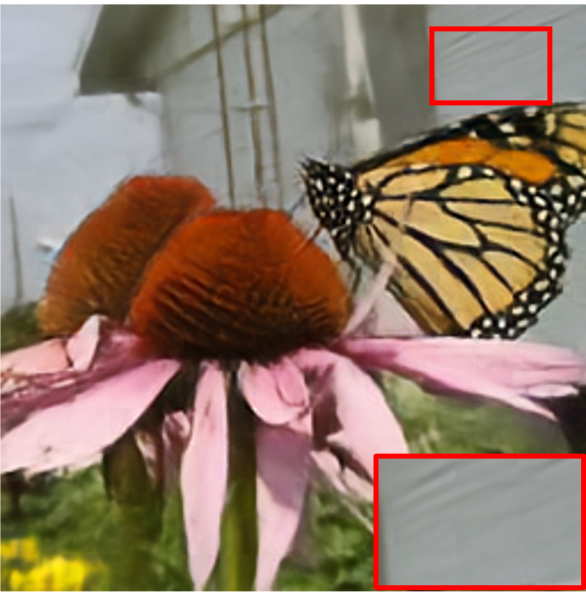}} &
    \raisebox{-0.5\height}{\includegraphics[width=\linewidth]{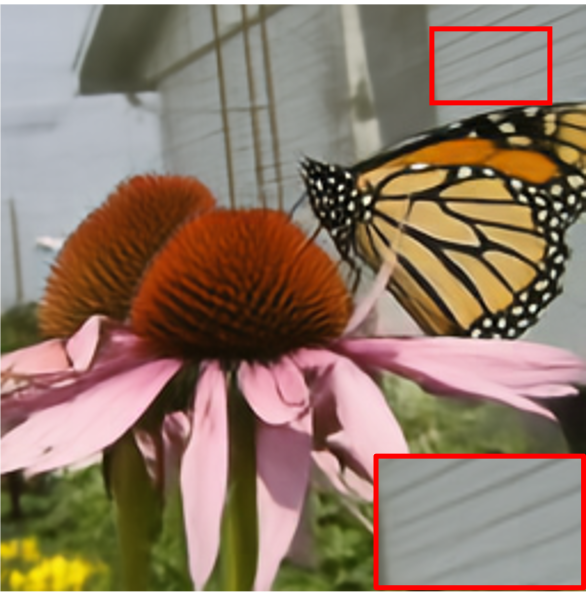}} &
    \raisebox{-0.5\height}{\includegraphics[width=\linewidth]{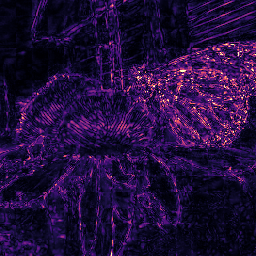}}
\end{tabular}

\caption{Visualization of intermediate and final predictions. We select Layer 6 of JiT-B/16 and visualize the noisy input, intermediate clean prediction, final clean prediction, and their residual at $t=0.3$ and $t=0.6$.}
\label{fig_intermediate_final_visualization}
\end{figure}
  
As shown in Fig.~\ref{fig_frequency_analysis}(a), the low-frequency band power of the intermediate predictions is already close to that of the final prediction at shallow layers (e.g.,  99.5$\%$ at Layer 4). 
In contrast, the high-frequency band power starts noticeably lower (87.7$\%$ at Layer 4) and increases with depth.
We further visualize the intermediate and final predictions in Fig.~\ref{fig_intermediate_final_visualization}.
The intermediate prediction already contains coarse, low-frequency image structure, while the final prediction refines local texture details.
The residual maps highlight these refinements around fine-scale structures.
These observations suggest that the intermediate prediction provides a naturally weak and lower-frequency counterpart to the final prediction.
Moreover, we compare the radial power of images generated by JiT with that of real images from the ImageNet validation set in Fig.~\ref{fig_frequency_analysis}(b) and observe a clear gap between them in the high-frequency band.
We therefore treat the intermediate and final predictions as the weak and strong predictions, respectively, and compute the guided clean prediction as
\begin{equation}
    \mathbf{x}_\mathrm{guided} = \mathbf{x}_\mathrm{inter} + w(\mathbf{x}_\mathrm{final}-\mathbf{x}_\mathrm{inter}).
    \label{eq_ssg_guidance}
\end{equation}
We refer to this as \textbf{self-guidance}. By Eq.~\ref{eq_x_to_v}, this is equivalent to applying the same guidance form to the velocity predictions.
Unlike AG and IG, the weak and strong predictions in self-guidance are derived from the same frozen pretrained backbone, without requiring a separate weak model or joint model training.
When $w=1$, $\mathbf{x}_\mathrm{guided}$ recovers the original final output, while $w>1$ extrapolates the prediction away from the intermediate prediction.
This extrapolation preserves the components shared by the intermediate and final outputs while amplifying the refinements introduced by the final prediction, particularly the high-frequency details that the model struggles to fit.
Table~\ref{tab_adapter_training_data} shows that an adapter trained on real images already improves generation quality over the JiT baseline.

\begin{table*}[t!]
\centering
\small
\setlength{\tabcolsep}{6pt}
\renewcommand{\arraystretch}{1.2}

\begin{tabular*}{\textwidth}{
@{\extracolsep{\fill}}
l c c c c c c
@{}
}
\toprule
Model &
Param. &
Epochs &
FID $\downarrow$ &
IS $\uparrow$ &
Precision $\uparrow$ &
Recall $\uparrow$ \\
\midrule

\multicolumn{7}{c}{\textit{\textbf{Generators in Latent Space}}} \\
\midrule

\textcolor{black}{DiT-XL/2 \cite{peebles2023scalable}}
& \textcolor{black}{675M + 49M}
& \textcolor{black}{1400}
& \textcolor{black}{2.27}
& \textcolor{black}{278.2}
& \textcolor{black}{0.83}
& \textcolor{black}{0.57} \\

\textcolor{black}{SiT-XL/2 \cite{ma2024sit}}
& \textcolor{black}{675M + 49M}
& \textcolor{black}{1400}
& \textcolor{black}{2.06}
& \textcolor{black}{277.5}
& \textcolor{black}{0.83}
& \textcolor{black}{0.59} \\

\textcolor{black}{REPA-XL/2 \cite{yu2024representation}}
& \textcolor{black}{675M + 49M}
& \textcolor{black}{800}
& \textcolor{black}{1.42}
& \textcolor{black}{305.7}
& \textcolor{black}{0.80}
& \textcolor{black}{0.64} \\

\textcolor{black}{DDT-XL/2 \cite{wang2026ddt}}
& \textcolor{black}{675M + 49M}
& \textcolor{black}{400}
& \textcolor{black}{1.26}
& \textcolor{black}{310.6}
& \textcolor{black}{0.79}
& \textcolor{black}{0.65} \\

\textcolor{black}{RAE-XL/2 \cite{zheng2025diffusion}}
& \textcolor{black}{839M + 415M}
& \textcolor{black}{800}
& \textcolor{black}{1.13}
& \textcolor{black}{262.6}
& \textcolor{black}{0.78}
& \textcolor{black}{0.67} \\

\textcolor{black}{LightningDiT+IG \cite{zhou2026guiding}}
& \textcolor{black}{678M + 49M}
& \textcolor{black}{680}
& \textcolor{black}{1.19}
& \textcolor{black}{269.0}
& \textcolor{black}{0.79}
& \textcolor{black}{0.66} \\

\midrule
\multicolumn{7}{c}{\textit{\textbf{Generators in Pixel Space}}} \\
\midrule

ADM-G \cite{dhariwal2021diffusion}
& 554M & 400 & 4.59 & 186.7 & \textbf{0.82} & 0.52 \\

RIN \cite{jabri2022scalable}
& 410M & 480 & 3.42 & 182.0 & -- & -- \\

VDM++ \cite{kingma2023understanding}
& 2B & -- & 2.12 & 267.7 & -- & -- \\

FractalMAR-H \cite{li2025fractal}
& 848M & 600 & 6.15 & \textbf{348.9} & 0.81 & 0.46 \\

PixelFlow-XL/4 \cite{chen2025pixelflow}
& 677M & 320 & 1.98 & 282.1 & 0.81 & 0.60 \\

PixNerd-XL/16 \cite{wang2025pixnerd}
& 700M & 320 & 1.93 & 297.0 & 0.79 & 0.59 \\

EPG-G/16 \cite{lei2025there}
& 1391M & 600 & 1.75 & 275.1 & 0.80 & 0.62 \\

PixelDiT-XL \cite{yu2026pixeldit}
& 797M & 320 & 1.61 & 292.7 & 0.78 & \textbf{0.64} \\

DiP-XL/16 \cite{chen2026dip}
& 631M & 600 & 1.79 & 281.9 & 0.80 & 0.62 \\

PixelGen-XL/16 \cite{ma2026pixelgen}
& 676M & 160 & 1.83 & 293.6 & 0.79 & 0.63 \\

PixelU-H/16 \cite{guo2026pixelu}
& 1168M & 600 & 1.63 & 305.9 & 0.79 & \textbf{0.64} \\

JiT-H/16 \cite{li2026back}
& 953M & 600 & 1.86 & 303.4 & 0.78 & 0.62 \\

PixelREPA-H/16 \cite{shin2026representation}
& 953M & 600 & 1.81 & 317.2 & 0.79 & 0.63 \\

DeCo-XL/16 \cite{ma2026deco}
& 682M & 600 & 1.69 & 304.0 & 0.79 & 0.63 \\

\midrule

\textbf{JiT-H/16 + SSG}
& \textcolor{black}{953M}$^\star$+ 63M
& 50\textsuperscript{\(\dagger\)} & 1.67 & 290.3 & 0.79 & 0.63 \\

\textbf{PixelREPA-H/16 + SSG}
& \textcolor{black}{953M}$^\star$+ 63M
& 50\textsuperscript{\(\dagger\)} & \textbf{1.59} & 304.2 & 0.79 & 0.63 \\

\textbf{DeCo-XL/16 + SSG}
& \textcolor{black}{682M}$^\star$+ 57M
& 50\textsuperscript{\(\dagger\)} & 1.63 & 303.5 & 0.79 & 0.63 \\
\bottomrule
\end{tabular*}

\caption{
Comparison of class-conditional generation on ImageNet 256$\times$256 with CFG. 
In our \textbf{SSG} rows, $\star$ denotes the frozen pretrained backbone parameters, which are not updated.
\(\dagger\) denotes adapter training epochs using 1M synthetic samples. 
For all methods, we report the best FID over guidance scales.
The best results among pixel-space models are highlighted in \textbf{bold}.}
\label{tab_quan_results_256}
\end{table*}
\paragraph{Training the Adapter with Synthetic Samples.} 
The adapter used in the previous analysis is trained on real images.
However, the role of the adapter in self-guidance is not to reproduce the full frequency components of real images.
Instead, it only needs to provide a coarse, low-frequency counterpart to the final prediction.
This motivates us to train the adapter on samples generated by the pretrained model itself.
As shown in Fig.~\ref{fig_ssg_method}(a), we use the pretrained diffusion model to generate a synthetic dataset.
Each generated image, paired with its sampling condition, is treated as a clean target for the adapter. 
We then train the adapter on this synthetic dataset using the objective in Eq.~\ref{eq_fm}, while keeping the backbone frozen (Fig.~\ref{fig_ssg_method}(b)).
During sampling, we apply the same guidance strategy defined in Eq.~\ref{eq_ssg_guidance}, as illustrated in Fig.~\ref{fig_ssg_method}(c).
We refer to this complete method as \textbf{Synthetic Self-Guidance} (SSG).
Surprisingly, Table~\ref{tab_adapter_training_data} shows that training the adapter with synthetic samples yields better generation quality than training it with real images.
As shown in Fig.~\ref{fig_frequency_analysis}(b), guidance with the synthetic-trained adapter brings the radial power spectrum of the generated images closest to that of real images, especially in the high-frequency band.
Importantly, SSG requires no real images and updates only the adapter, using less than 1$\%$ of full-model training compute.
This makes SSG a plug-in method for improving pretrained pixel diffusion models.
In practice, SSG can be used either alone or together with CFG.
When combined with CFG, we first apply self-guidance independently to the conditional and unconditional clean predictions, convert both guided predictions into velocities, and then apply CFG.

\section{Experiments}
\subsection{Experiment Settings}
We evaluate SSG on all JiT \cite{li2026back} variants for class-conditional ImageNet \cite{russakovsky2015imagenet} generation 
at 256 and 512 resolutions, and further on PixelREPA-H \cite{shin2026representation} and DeCo-XL \cite{ma2026deco}.
For all models, we apply SSG to the officially released pretrained checkpoints while keeping their backbones frozen, and take the baseline results from the corresponding papers.
\paragraph{Adapter Settings.}
The JiT adapter consists of one transformer block \cite{vaswani2017attention} for the base and large models and two blocks for the huge models, followed by a linear layer that maps features back into pixel space.
PixelREPA-H adopts the same adapter configuration as JiT-H.
Since DeCo employs a hierarchical architecture, we attach the adapter to its intermediate encoder representation and use its pixel decoder to produce the intermediate prediction.
The attachment layers are provided in the supplementary material.
\paragraph{Synthetic Data.}
Each pretrained model generates its own synthetic dataset using the best CFG setting in its paper.
We generate 1K samples per class (1M images in total), approximately matching the scale of the 1.28M ImageNet training images used in the real-data comparison.
We show in Table~\ref{tab_synthetic_dataset_size} that fewer synthetic samples suffice in practice.

\paragraph{Training and Sampling.}
We train the adapters of JiT-H, PixelREPA-H, and DeCo-XL for 50 epochs and those of JiT-B and JiT-L for 30 epochs.
In all cases, adapter training requires less than 1$\%$ of the computational cost of training the corresponding baseline model from scratch.
Following JiT, we use the Heun sampler with 50 steps and evaluate SSG both with and without CFG.
Without CFG, SSG is applied throughout the sampling process.
When combined with CFG, both CFG and SSG are applied within the timestep interval $t\in[0.1,1.0]$.
We report the best result for each model by sweeping the CFG and SSG scales, with the specific scales provided in the supplementary material.

\paragraph{Evaluation.}
We evaluate generation performance on the ImageNet validation set and follow the JiT evaluation protocol by generating 50K class-balanced samples.
We report FID \cite{heusel2017gans}, Inception Score (IS) \cite{salimans2016improved}, Precision and Recall \cite{kynkaanniemi2019improved}, and compare SSG with the most recent pixel diffusion models.

\subsection{Main Results}
\label{sec_exp}
\paragraph{Class-Conditional Image Generation.}
Tables~\ref{tab_quan_results_256} and \ref{tab_quan_results_512} report results with CFG at 256 and 512 resolutions.
SSG improves FID across all backbones: from 1.86 to 1.67 (10.2$\%$) for JiT-H/16, from 1.94 to 1.84 (5.2$\%$) for JiT-H/32, and from 1.81 to 1.59 (12.2$\%$) for PixelREPA-H/16.
It also transfers to hierarchical architectures, improving DeCo-XL/16 from 1.69 to 1.63 without per-layer tuning.
IS decreases slightly because we select the CFG and SSG scales for the lowest FID, for which SSG typically favors a lower CFG scale.
The fixed-CFG comparison in Fig.~\ref{fig_adapter_layer_and_guidance_scales}(b) further shows that SSG still improves FID at the optimal CFG scale of the original baseline.
Without CFG (Table~\ref{tab_quan_results_256_without_cfg}), the gains are substantially larger: SSG reduces FID by more than 50$\%$ for all models.

\begin{table}[t]
\centering
\small
\setlength{\tabcolsep}{2.0pt}
\renewcommand{\arraystretch}{1.12}

\begin{tabular}{
@{}
l c c c c c c
@{}
}
\toprule
Model &
Params &
Ep. &
FID $\downarrow$ &
IS $\uparrow$ &
Prec. $\uparrow$ &
Rec. $\uparrow$ \\
\midrule

\multicolumn{7}{c}{\textit{\textbf{Generators in Latent Space}}} \\
\midrule

\textcolor{black}{DiT-XL/2}
& \textcolor{black}{675M + 49M}
& \textcolor{black}{600}
& \textcolor{black}{3.04}
& \textcolor{black}{240.8}
& \textcolor{black}{0.84}
& \textcolor{black}{0.54} \\

\textcolor{black}{SiT-XL/2}
& \textcolor{black}{675M + 49M}
& \textcolor{black}{600}
& \textcolor{black}{2.62}
& \textcolor{black}{252.2}
& \textcolor{black}{0.84}
& \textcolor{black}{0.57} \\

\textcolor{black}{REPA-XL/2}
& \textcolor{black}{675M + 49M}
& \textcolor{black}{800}
& \textcolor{black}{2.08}
& \textcolor{black}{274.6}
& \textcolor{black}{0.83}
& \textcolor{black}{0.58} \\

\midrule
\multicolumn{7}{c}{\textit{\textbf{Generators in Pixel Space}}} \\
\midrule

ADM-G
& 554M & 400 & 7.72 & 172.7 & \textbf{0.84} & 0.53 \\

PixNerd-XL/16
& 700M & 340 & 2.84 & 245.6 & 0.80 & 0.59 \\

EPG-L/32
& 540M & 800 & 2.35 & 295.4 & 0.82 & 0.57 \\

DeCo-XL/16
& 682M & 340 & 2.22 & 290.0 & 0.80 & 0.60 \\

PixelU-H/32
& 1152M & 600 & 1.92 & \textbf{322.1} & 0.80 & 0.58 \\

JiT-H/32
& 956M & 600 & 1.94 & 309.1 & 0.80 & 0.61 \\

\midrule

\textbf{JiT-H/32+SSG}
& \textcolor{black}{956M}$^\star$+ 66M
& 50\textsuperscript{\(\dagger\)}
& \textbf{1.84}
& 305.2
& 0.80
& \textbf{0.63} \\

\bottomrule
\end{tabular}

\caption{
Comparison of class-conditional generation on ImageNet 512$\times$512 with CFG. Same settings as Table \ref{tab_quan_results_256}. 
}
\label{tab_quan_results_512}
\end{table}

\subsection{Ablation Study}
\paragraph{Synthetic Samples Outperform Real Images for Adapter Training.}
Table~\ref{tab_adapter_training_data} compares adapters trained on real and synthetic images.
Training the adapter on real images consistently improves the JiT baselines, validating the effectiveness of
self-guidance.
The adapter trained on synthetic samples outperforms the real-trained one across nearly all settings.
For JiT-H/16, synthetic training reduces FID from 2.51 to 2.26 without CFG and from 1.78 to 1.67 with CFG.
Fig.~\ref{fig_frequency_analysis}(b) supports this finding: the adapter trained on the synthetic dataset produces samples whose radial power is closer to that of real images than the real-trained adapter.

\paragraph{Fewer Synthetic Samples Suffice.}
Table~\ref{tab_synthetic_dataset_size} studies the effect of synthetic dataset size.
Reducing the dataset from 1M to 10K samples (a 100$\times$ reduction) changes FID by less than 0.05 for both JiT-B/16 (3.29 to 3.31) and JiT-B/32 (3.67 to 3.69).
Note that we vary only the dataset size while keeping the total number of training iterations fixed.
In our main experiments, we use 1M samples by default, approximately matching the scale of the ImageNet training set for a fair comparison with real-trained adapters.

% \paragraph{Comparison with Jointly Trained IG.}
% Table~\ref{tab_ig_comparison} compares SSG with IG \cite{zhou2026guiding}, which jointly trains JiT backbone and the intermediate head for 600 epochs.
% %
% Under the same JiT-B/16 configuration, jointly trained IG achieves 3.41 FID, outperforming the real-trained adapter (3.50) but remaining behind SSG (3.29).
% %
% Thus, in this setting, SSG provides stronger guidance while updating only the adapter of a frozen pretrained model.

%   \begin{table}[t]
%   \centering
%   \small
%   \setlength{\tabcolsep}{4pt}
%   \renewcommand{\arraystretch}{1.1}

%   \begin{tabular}{
%   @{}
%   l c c c
%   @{}
%   }
%   \toprule
%   Method &
%   Backbone &
%   Training Data &
%   FID $\downarrow$ \\
%   \midrule

%   JiT-B/16 baseline
%   & -- & -- & 3.66 \\

%   Real-trained adapter
%   & Frozen & Real & 3.50 \\

%   Jointly trained IG
%   & Updated & Real & 3.41 \\

%   \textbf{SSG}
%   & Frozen & Synthetic & \textbf{3.29} \\

%   \bottomrule
%   \end{tabular}

%   \caption{
%   Comparison with jointly trained IG on JiT-B/16.
%   All guidance methods use the same intermediate layer, head architecture,
%   and sampling settings.
%   }
%   \label{tab_ig_comparison}
%   \end{table}

\begin{figure}[t!]
\centering

\begin{subfigure}[t]{0.49\linewidth}
\centering
\includegraphics[width=\linewidth]{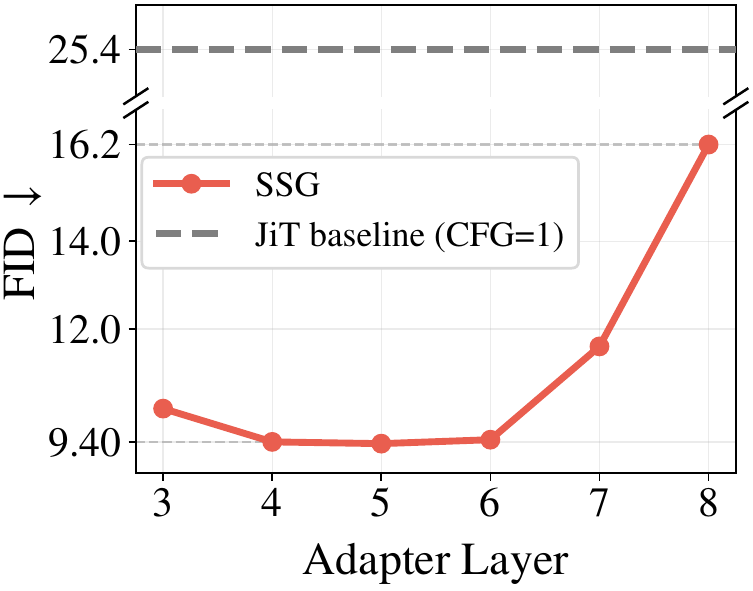}
\caption{Adapter placement}
\label{fig:adapter_layer}
\end{subfigure}
\hfill
\begin{subfigure}[t]{0.49\linewidth}
\centering
\includegraphics[width=\linewidth]{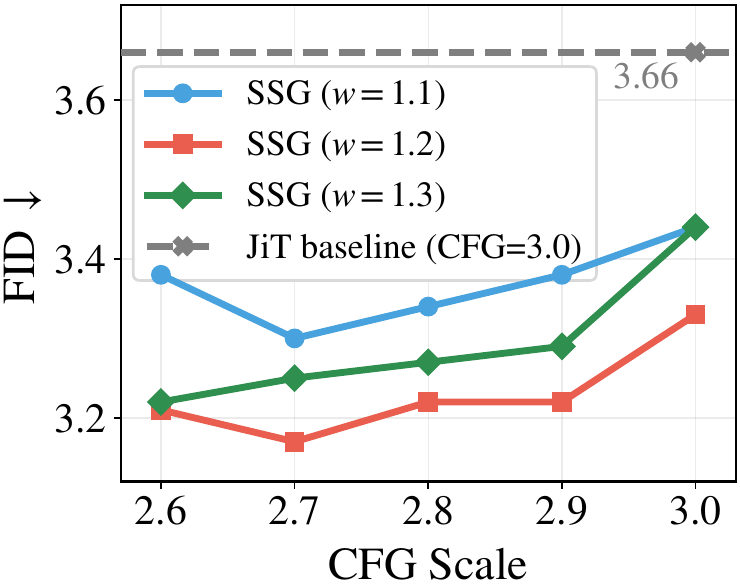}
\caption{Guidance-scale sweep}
\label{fig:guidance_scales}
\end{subfigure}

\caption{
Ablations on adapter placement and guidance scales using JiT-B/16.
(a) FID for adapters attached to different layers without CFG.
(b) FID across different CFG and SSG scales.
Dashed lines indicate the corresponding JiT baselines.
}
\label{fig_adapter_layer_and_guidance_scales}
\end{figure}

\paragraph{Early-to-Middle Layers Provide the Best Guidance.}
Fig.~\ref{fig_adapter_layer_and_guidance_scales}(a) evaluates attachment layers on JiT-B/16 without CFG.
Layers 4--6 achieve comparable FID, while Layer 3 is slightly worse and Layers 7--8 degrade sharply toward the baseline.
At deeper layers, the intermediate prediction approaches the final prediction, reducing their discrepancy and weakening self-guidance.
We also observe this trend for large and huge models.
In practice, we favor earlier layers within this range to reduce adapter-training compute.
The layer for each model is provided in the supplementary material.

\begin{table}[t]
\centering
\small
\setlength{\tabcolsep}{2.0pt}
\renewcommand{\arraystretch}{1.12}

\begin{tabular}{
@{}
l c c c c c c
@{}
}
\toprule
Model &
Params &
Ep. &
FID $\downarrow$ &
IS $\uparrow$ &
Prec. $\uparrow$ &
Rec. $\uparrow$ \\
% \midrule

% \multicolumn{7}{c}{\textit{\textbf{Latent-space Generators}}} \\
% \midrule

% \textcolor{gray}{DiT-XL/2}
% & \textcolor{gray}{675M+49M}
% & \textcolor{gray}{1400}
% & \textcolor{gray}{9.62}
% & \textcolor{gray}{121.5}
% & \textcolor{gray}{0.67}
% & \textcolor{gray}{0.67} \\

% \textcolor{gray}{SiT-XL/2}
% & \textcolor{gray}{675M+49M}
% & \textcolor{gray}{1400}
% & \textcolor{gray}{8.61}
% & \textcolor{gray}{131.7}
% & \textcolor{gray}{0.68}
% & \textcolor{gray}{0.67} \\

% \textcolor{gray}{REPA-XL/2}
% & \textcolor{gray}{675M+49M}
% & \textcolor{gray}{800}
% & \textcolor{gray}{5.90}
% & \textcolor{gray}{157.8}
% & \textcolor{gray}{0.70}
% & \textcolor{gray}{0.69} \\

% \textcolor{gray}{LightningDiT+IG \cite{zhou2026guiding}}
% & \textcolor{gray}{678M + 49M}
% & \textcolor{gray}{680}
% & \textcolor{gray}{1.34}
% & \textcolor{gray}{229.3}
% & \textcolor{gray}{0.79}
% & \textcolor{gray}{0.66} \\

% \midrule
% \multicolumn{7}{c}{\textit{\textbf{Pixel-space Generators}}} \\
\midrule

ADM-G
& 554M & 400 & 10.94 & -- & 0.69 & 0.63 \\

PixelFlow-XL
& 677M & 320 & 12.23 & 103.3 & 0.63 & 0.66 \\

PixNerd-XL
& 700M & 320 & 15.61 & 88.9 & 0.59 & \textbf{0.68} \\

DeCo-XL/16
& 682M & 320 & 14.88 & 88.2 & 0.60 & \textbf{0.68} \\

PixelGen-XL/16
& 676M & 80 & 5.11 & {159.2 }& {0.72 }& 0.63 \\

JiT-B/16 
& 131M & 600 & 25.42 & 63.2 & 0.54 & 0.66 \\

JiT-L/16 
& 458M & 600 & 13.85 & 104.2 & 0.62 & 0.67 \\

JiT-H/16 
& 953M & 600 &  7.15 & 151.7 & 0.68 & 0.67 \\

\midrule

\textbf{JiT-B/16+SSG} 
& \textcolor{black}{131M}$^\star$+ 12M 
& 30\textsuperscript{\(\dagger\)}  
& 9.47
& 115.5
& 0.70 
& 0.57
\\

\textbf{JiT-L/16+SSG} 
& \textcolor{black}{459M}$^\star$+ 22M 
& 30\textsuperscript{\(\dagger\)}  
& 4.47
& 174.5
& 0.76 
& 0.59
\\

\textbf{JiT-H/16+SSG} 
& \textcolor{black}{953M}$^\star$+ 63M 
& 50\textsuperscript{\(\dagger\)}  
& \textbf{2.26}
& \textbf{223.7}
&\textbf{0.78}
& 0.62
\\

\bottomrule
\end{tabular}

\caption{
Comparison of class-conditional generation on ImageNet at
$256\times256$ resolution without CFG.
All other settings and notation follow Table~\ref{tab_quan_results_256}.
}
\label{tab_quan_results_256_without_cfg}
\end{table}

\paragraph{Effect of Guidance Scales.}
Fig.~\ref{fig_adapter_layer_and_guidance_scales}(b) sweeps the SSG scale $w$ and CFG scale on JiT-B/16.
SSG improves the baseline across a wide range of scale combinations.
At the baseline's optimal CFG scale of 3.0, SSG with $w=1.2$ reduces FID from 3.66 to 3.33, showing that the gain is not merely due to the larger guidance
search space.
The guidance scales of each configuration are provided in the supplementary material.

\begin{table}[t]
\centering
\small
\setlength{\tabcolsep}{2.0pt}
\renewcommand{\arraystretch}{1.12}

\begin{tabularx}{\columnwidth}{
@{}
l
*{6}{>{\centering\arraybackslash}X}
@{}
}
\toprule
&
\multicolumn{3}{c}{W/o CFG} &
\multicolumn{3}{c}{With CFG} \\
\cmidrule(lr){2-4}
\cmidrule(lr){5-7}

Model &
Baseline &
Real &
Syn. &
Baseline &
Real &
Syn. \\
\midrule

JiT-B/16
& 25.42
& 9.73
& \textbf{9.47}
& 3.66
& 3.50
& \textbf{3.29} \\

JiT-B/32
& 28.57
& 14.16
& \textbf{13.37}
& 4.02
& 3.91
& \textbf{3.67} \\

JiT-L/16
& 13.85
& 4.72
& \textbf{4.47}
& 2.36
& 2.34
& \textbf{2.23} \\

JiT-L/32
& 13.91
& 4.86
& \textbf{4.51}
& 2.53
& 2.42
& \textbf{2.31} \\

JiT-H/16
& 7.15
& 2.51
& \textbf{2.26}
& 1.86
& 1.78
& \textbf{1.67} \\

JiT-H/32
& 7.32
& \textbf{2.66}
& 2.73
& 1.94
& 1.91
& \textbf{1.84} \\

\bottomrule
\end{tabularx}

\caption{
Ablation on adapter training data across JiT variants. 
All entries are FID ($\downarrow$).
Real denotes adapters trained on the ImageNet training set (1.28M images), while Syn.\ denotes adapters trained on 1M synthetic samples.
For each variant, the Real and Syn.\ adapters use identical settings.
}
\label{tab_adapter_training_data}
\end{table}

\begin{table}[t]
\centering
\small
\setlength{\tabcolsep}{5pt}
\renewcommand{\arraystretch}{1.12}

\begin{tabular}{
@{}
c
cc
cc
@{}
}
\toprule
\multirow{2}{*}{Samples} &
\multicolumn{2}{c}{JiT-B/16} &
\multicolumn{2}{c}{JiT-B/32} \\
\cmidrule(lr){2-3}
\cmidrule(lr){4-5}
&
FID $\downarrow$ &
IS $\uparrow$ &
FID $\downarrow$ &
IS $\uparrow$ \\
\midrule

1M
& 3.29
& \textbf{274.0}
& \textbf{3.67}
& 272.7 \\

100K
& \textbf{3.25}
& 270.4
& --
& -- \\

10K
& 3.31
& 268.9
& 3.69
& \textbf{274.8} \\

\bottomrule
\end{tabular}

\caption{
Ablation on synthetic dataset size.
All adapters are trained with the same number of iterations.
}
\label{tab_synthetic_dataset_size}
\end{table}

\begin{table}[t]
\centering
\small
\setlength{\tabcolsep}{1.5pt}
\renewcommand{\arraystretch}{1.12}

\begin{tabularx}{\columnwidth}{
@{}
l
>{\centering\arraybackslash}X
>{\centering\arraybackslash}X
>{\centering\arraybackslash}X
>{\centering\arraybackslash}X
@{}
}
\toprule
Adapter &
Params &
Rel. FLOPs &
FID w/o CFG $\downarrow$ &
FID with CFG $\downarrow$ \\
\midrule

\multicolumn{5}{c}{\textbf{JiT-B/16, Layer 6}} \\
\midrule

No adapter
& 0
& 0
& 25.42
& 3.66 \\

Linear only
& 1.77M
& 0.65\%
& 36.44
& 7.95 \\

% \rowcolor[gray]{0.94}
1 block (default)
& 12.40M
& 0.99\%
& 9.47
& 3.29 \\

2 blocks
& 23.03M
& 1.33\%
& 9.14
& 3.22 \\

\midrule
\multicolumn{5}{c}{\textbf{JiT-H/16, Layer 8}} \\
\midrule

No adapter
& 0
& 0
& 7.15
& 1.86 \\

1 block
& 33.78M
& 0.69\%
& 2.51
& 1.76 \\

% \rowcolor[gray]{0.94}
2 blocks (default)
& 63.29M
& 0.88\%
& 2.26
& 1.67 \\

\bottomrule
\end{tabularx}

\caption{
Ablation on adapter capacity.
Rel.\ FLOPs denotes the FLOPs for adapter training relative to training the corresponding full JiT model for 600 epochs.
All other training and sampling settings across different variants are kept fixed.
}
\label{tab_adapter_capacity}
\end{table}

\paragraph{A Lightweight Adapter Suffices.}
Table~\ref{tab_adapter_capacity} studies the effect of adapter capacity.
We fix the SSG and CFG scales across adapter variants.
A linear layer performs worse than the JiT baseline (36.44 vs.\ 25.42 without CFG) and introduces patch-boundary artifacts.
Adding a transformer block removes these artifacts and improves generation quality.
On JiT-B/16, two blocks slightly improve FID but raise adapter-training compute above $1\%$ of full-model training, so we use one block for JiT-B and JiT-L.
For JiT-H, two blocks provide a clearer gain at low cost, so we adopt this setting.
\begin{table}[t]
\centering
\small
\setlength{\tabcolsep}{4pt}
\renewcommand{\arraystretch}{1.12}

\begin{tabular}{
@{}
l
*{3}{>{\centering\arraybackslash}m{0.11\columnwidth}}
@{}
}
\toprule
&
\multicolumn{3}{c}{Generation CFG} \\
\cmidrule(lr){2-4}

Model &
1.0 &
2.0 &
3.0 \\
\midrule

JiT-B/16
& 3.25
& 3.31
& 3.29 \\

JiT-B/32
& 3.72
& 3.71
& 3.67 \\

\bottomrule
\end{tabular}

\caption{
Effect of the CFG scale used to generate synthetic training samples.
All entries are FID ($\downarrow$).
During evaluation, we fix the CFG scale to 2.9 for both models and set the SSG scale to 1.3 and 1.2 for JiT-B/16 and JiT-B/32, respectively.
}
\label{tab_synthetic_generation_cfg}
\end{table}

\paragraph{SSG Is Robust to the Generation CFG.}
By default, we generate the synthetic data using the best CFG scale reported for each pretrained model.
Table~\ref{tab_synthetic_generation_cfg} varies this generation CFG and shows
that the downstream FID is nearly unchanged: it stays within 0.06 on
JiT-B/16 (3.25 to 3.31) and 0.05 on JiT-B/32 (3.67 to 3.72).
This indicates that SSG does not require a carefully chosen generation CFG, and we simply adopt the reported best CFG without additional tuning.

\paragraph{Comparison with IG.}
We implement IG \cite{zhou2026guiding} on JiT by jointly training the backbone and an intermediate head at Layer 6 for 600 epochs, matching the JiT-B/16 baseline schedule.
As shown in Table~\ref{tab_ig_comparison}, it reaches 3.41 FID, outperforming the frozen adapter trained on real images (3.50) but remaining behind SSG (3.29).

  \begin{table}[t!]
  \centering
  \small
  \setlength{\tabcolsep}{4pt}
  \renewcommand{\arraystretch}{1.1}

  \begin{tabular}{
  @{}
  l c c c
  @{}
  }
  \toprule
  Method &
  Backbone &
  Training Data &
  FID $\downarrow$ \\
  \midrule

  JiT-B/16
  & -- & -- & 3.66 \\

  Adapter+real
  & Frozen & Real & 3.50 \\

  JiT+IG
  & Updated & Real & 3.41 \\

  \textbf{SSG}
  & Frozen & Synthetic & \textbf{3.29} \\

  \bottomrule
  \end{tabular}

  \caption{
  Comparison with jointly trained IG on JiT-B/16.
We implement IG following \cite{zhou2026guiding}; all methods use Layer 6 as the intermediate layer.
  }
  \label{tab_ig_comparison}
  \end{table}

\section{Conclusion}
We presented \textbf{Synthetic Self-Guidance} (SSG), a plug-in method for improving pretrained pixel diffusion models while keeping their backbones frozen.
SSG decodes an intermediate representation into a coarse prediction and uses its discrepancy from the final prediction as self-guidance during sampling.
The adapter is trained solely on model-generated samples, which outperform real images for this purpose in nearly all evaluated settings.
Experiments across multiple pixel diffusion models show consistent FID improvements with low adapter-training cost.

\section*{Supplementary Material}

  \appendix
  \setcounter{secnumdepth}{1}

  \setcounter{figure}{0}
  \renewcommand{\thefigure}{S\arabic{figure}}

  \setcounter{table}{0}
  \renewcommand{\thetable}{S\arabic{table}}

  \setcounter{equation}{0}
  \renewcommand{\theequation}{S\arabic{equation}}

  \section{Implementation Details}

Table~\ref{tab_supp_implementation} summarizes the backbone, adapter, training, and sampling configurations used in our experiments.

\paragraph{Adapter Initialization.}
For JiT and PixelREPA, the adapter blocks are initialized from the late backbone blocks listed in Table~\ref{tab_supp_implementation}, while the output layer is initialized from the pretrained final prediction head.
For DeCo, we initialize the adapter from its last two encoder blocks and copy the pretrained pixel decoder as the adapter output layer.
During adapter training, we update only the adapter.

\paragraph{Training Details.}
Following JiT, we use Adam with $\beta_1=0.9$ and $\beta_2=0.95$, without weight decay or learning-rate warmup.
For JiT and PixelREPA, a base learning rate of 5e-5 is scaled to 2e-4 with a global batch size of 1024.
DeCo uses a learning rate of 5e-5 and a global batch size of 256.
JiT and PixelREPA train the adapter to predict the clean image, whereas DeCo directly optimizes the velocity prediction.
No representation supervision or auxiliary frequency loss is used for adapter training.

\paragraph{Sampling Details.}
All models use the Heun sampler with 50 steps.
JiT models at 256 and 512 resolutions use noise scales of 1.0 and 2.0, respectively, while PixelREPA and DeCo use a noise scale of 1.0.
Without CFG, SSG is applied throughout sampling; when combined with CFG, both guidance methods are applied within $t\in[0.1,1.0]$.
The CFG and SSG scales used for each model and resolution are provided in Table~\ref{tab_supp_guidance_scales}.

\section{Guidance-Scale Results}

\begin{figure}[t!]
    \centering
    \begin{subfigure}[t]{0.49\linewidth}
      \centering
      \includegraphics[width=\linewidth]
      {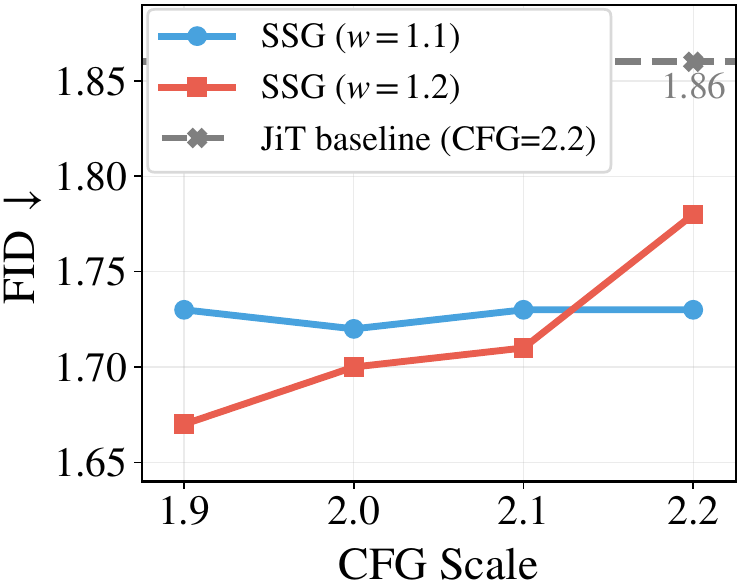}
      \caption{JiT-H/16}
      \label{fig_supp_guidance_h16}
    \end{subfigure}
    \hfill
    \begin{subfigure}[t]{0.49\linewidth}
      \centering
      \includegraphics[width=\linewidth]
      {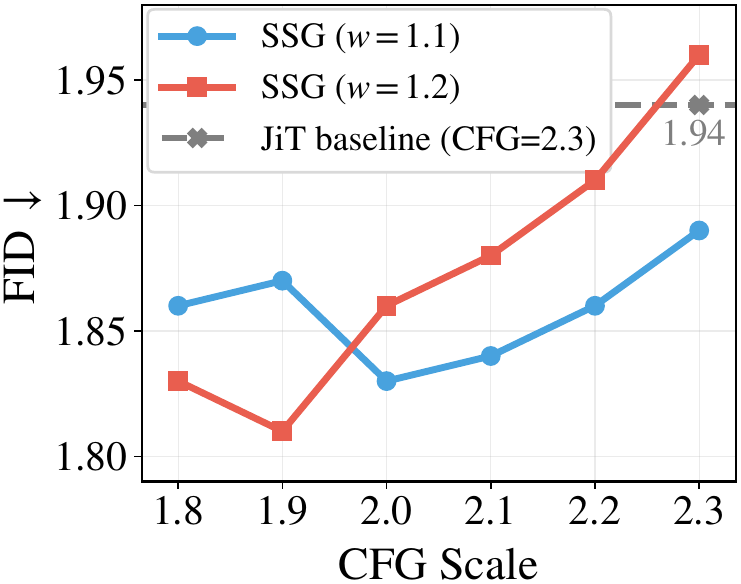}
      \caption{JiT-H/32}
      \label{fig_supp_guidance_h32}
    \end{subfigure}

    \caption{
    Guidance-scale sweeps on JiT-H/16 and JiT-H/32.
    Dashed lines indicate the corresponding JiT baselines.
    }
    \label{fig_supp_guidance_sweeps}
  \end{figure}

Fig.~\ref{fig_supp_guidance_sweeps} presents the full guidance-scale sweeps for JiT-H/16 and JiT-H/32.
SSG improves FID over a broad range of CFG and SSG scales, showing that its effectiveness is not sensitive to a specific scale combination.
At the CFG scales used by the original baselines, setting $w=1.1$ reduces FID from 1.86 to 1.73 on JiT-H/16 and from 1.94 to 1.89 on JiT-H/32.
This fixed-CFG comparison confirms that the improvement does not merely result from searching over additional guidance-scale combinations.

\section{Frequency-Domain Analysis Details}
\paragraph{Layer-Wise Band Power.}
For the layer-wise analysis, we use 512 images from the ImageNet validation set and construct noisy inputs at $t=0.5$.
We obtain intermediate clean predictions from independently trained adapters attached to Layers 4, 6, and 8 of JiT-B/16, together with the final clean prediction.
We apply an orthonormal two-dimensional Fourier transform to each RGB channel of the predictions and sum the squared magnitudes across channels.
Frequency coefficients are grouped into one-pixel radial bands, and their power is averaged across images.
We exclude the DC component and normalize the radial frequency by the Nyquist frequency.
The low- and high-frequency powers are computed over $(0,0.5)$ and $[0.5,1]$, respectively, and normalized by the corresponding band power of the final prediction.

\begin{table}[t!]
  \centering
  {\small
  \setlength{\tabcolsep}{4pt}
  \renewcommand{\arraystretch}{1.15}
  \begin{tabular}{@{}lccc@{}}
  \toprule
  &
  W/o CFG &
  \multicolumn{2}{c}{With CFG} \\
  \cmidrule(lr){2-2}
  \cmidrule(lr){3-4}

  Model &
  SSG  &
  CFG &
  SSG  \\
  \midrule

  JiT-B/16
  & 1.7 & 2.9 & 1.2 \\

  JiT-B/32
  & 1.7 & 2.9 & 1.3 \\

  JiT-L/16
  & 1.4 & 2.2 & 1.1 \\

  JiT-L/32
  & 1.4 & 2.3 & 1.1 \\

  JiT-H/16
  & 1.4 & 1.9 & 1.2 \\

  JiT-H/32
  & 1.3 & 2.1 & 1.1 \\

  PixelREPA-H/16
  & -- & 2.0 & 1.2 \\

  DeCo-XL/16
  & -- & 2.9 & 1.05 \\

  \bottomrule
  \end{tabular}
  }
  \caption{
  Specific guidance scales used for evaluation.
  Other sampling settings are provided in Table~\ref{tab_supp_implementation}.
  }
  \label{tab_supp_guidance_scales}
  \end{table}

\paragraph{Generated-Image Radial Power.}
We compare 2048 class-balanced ImageNet validation images with the same number of samples from the JiT-B/16 baseline and self-guidance using adapters trained on real or synthetic images.
All generated results use the same class labels and initial noise.
%
% We use Euler sampling with 50 steps without CFG and apply self-guidance throughout sampling.
%
Real and generated images are clamped in $[0,1]$, and their radial power is computed using the same orthonormal Fourier transform and channel aggregation described above.
We average the two-dimensional power maps over all images and exclude the DC component.
The remaining frequency coefficients are grouped into radial bands according to their distance from the zero-frequency location, and the power is summed within each band.
We normalize the radial frequency such that 1 corresponds to the Nyquist frequency.
For visualization, we show normalized frequencies from 0.12 to 0.96 on a logarithmic scale and divide all curves by the same constant, preserving their relative differences.

% \paragraph{Residual Visualization.}
% The residual maps visualize the RGB root-mean-square difference between the raw intermediate and final clean predictions:
% \begin{equation}
% R(i,j)
% =
% \sqrt{
% \frac{1}{3}
% \sum_{c=1}^{3}
% \left(
% x_{\mathrm{final}}^{c}(i,j)
% -
% x_{\mathrm{inter}}^{c}(i,j)
% \right)^2
% }.
% \end{equation}
% %
% The residual maps at $t=0.3$ and $t=0.6$ use the same color scale, whose maximum is set to their joint 99th percentile.

\section{Additional Qualitative Results}
\begin{figure*}[p]
\centering

\begin{minipage}[t]{0.495\textwidth}
\centering
\includegraphics[width=\linewidth]{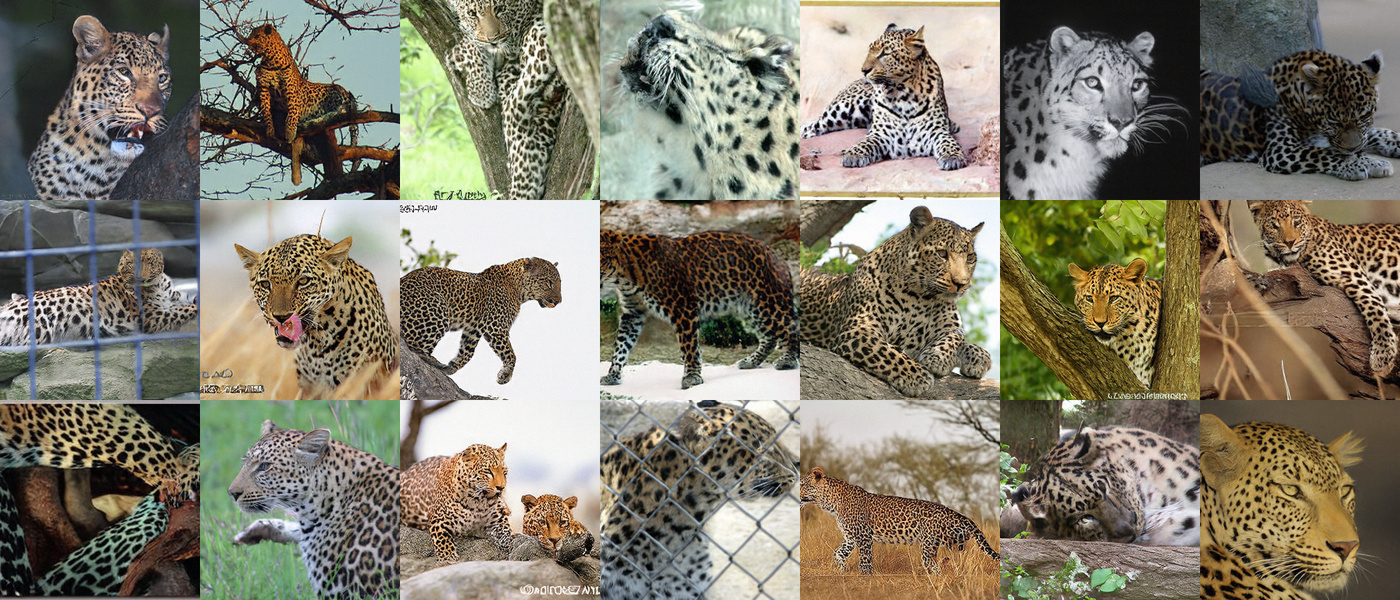}
\par\vspace{-0.25em}
{\scriptsize class 288: leopard, Panthera pardus\par}
\end{minipage}\hfill
\begin{minipage}[t]{0.495\textwidth}
\centering
\includegraphics[width=\linewidth]{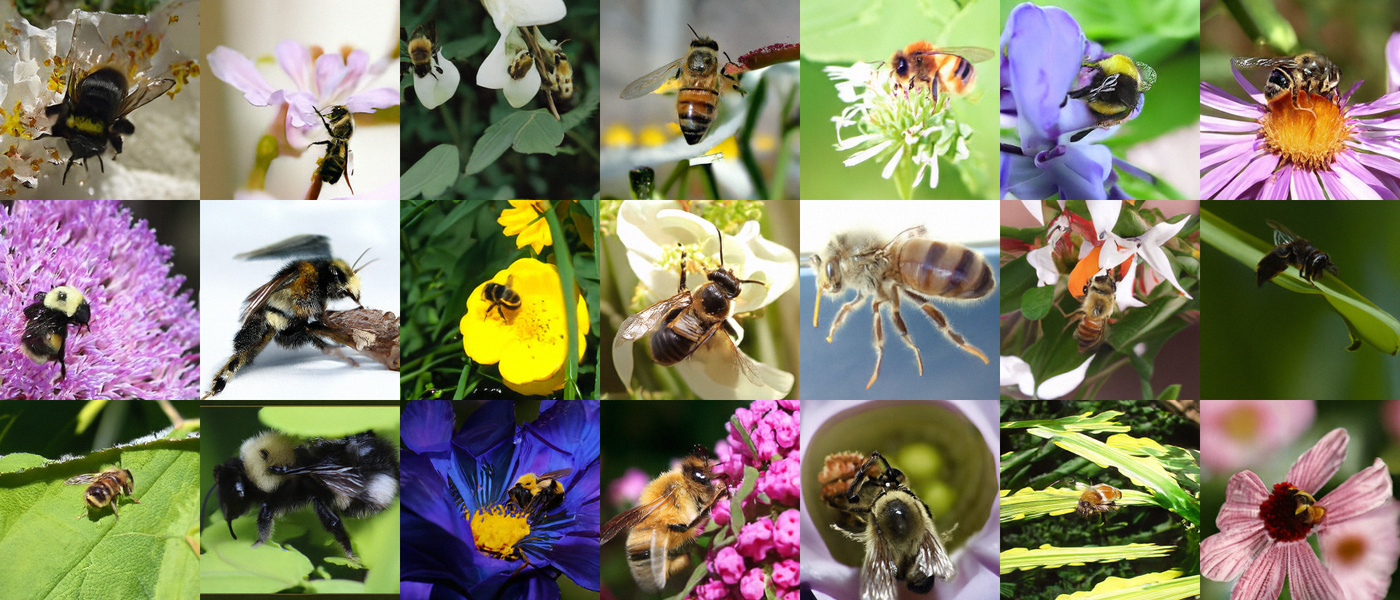}
\par\vspace{-0.25em}
{\scriptsize class 309: bee\par}
\end{minipage}

\vspace{0.15em}

\begin{minipage}[t]{0.495\textwidth}
\centering
\includegraphics[width=\linewidth]{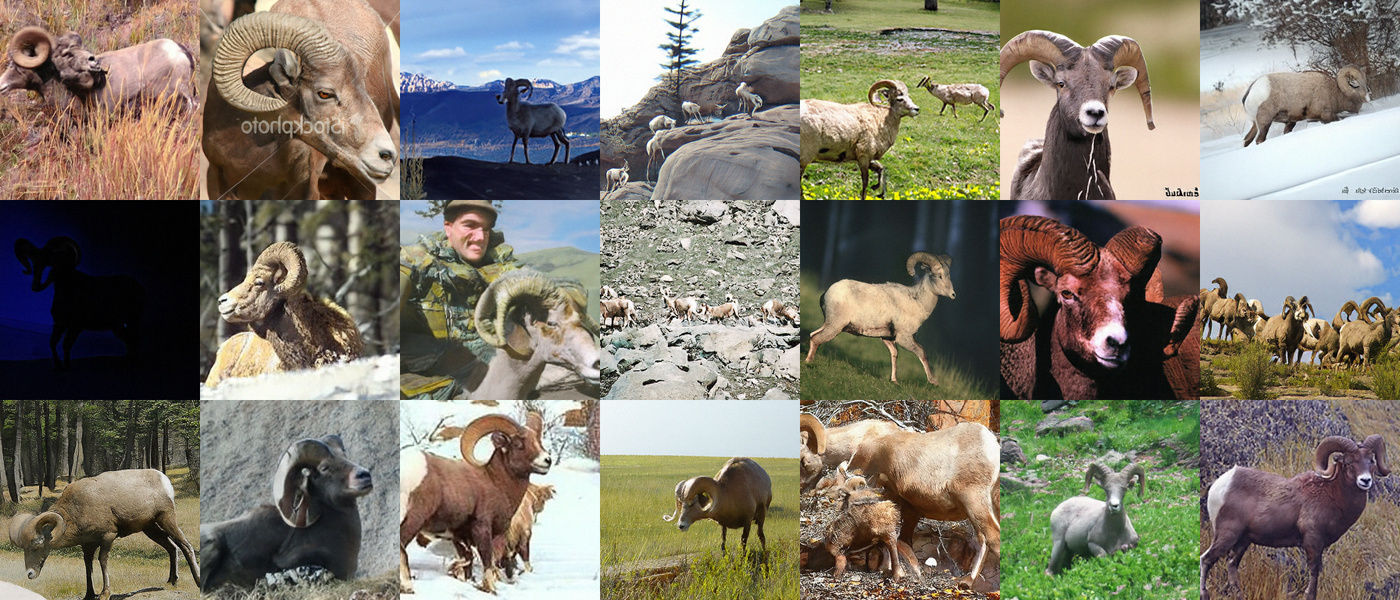}
\par\vspace{-0.25em}
{\scriptsize class 349: bighorn, bighorn sheep, cimarron, Rocky Mountain bighorn\par}
\end{minipage}\hfill
\begin{minipage}[t]{0.495\textwidth}
\centering
\includegraphics[width=\linewidth]{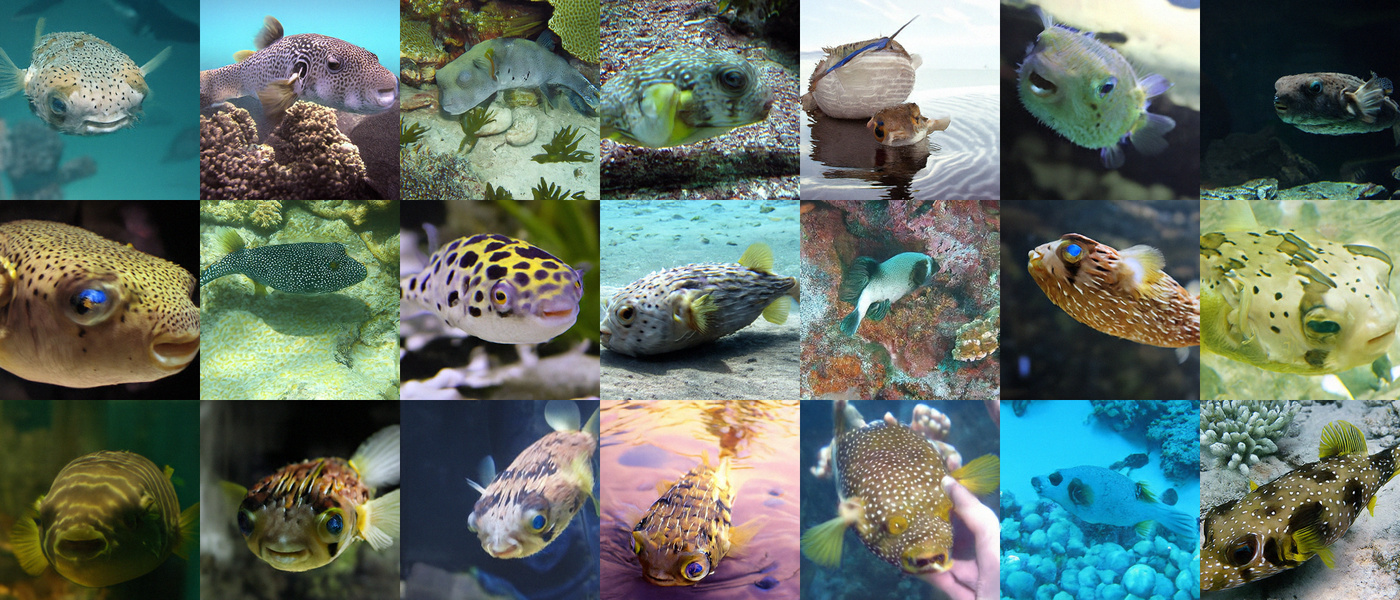}
\par\vspace{-0.25em}
{\scriptsize class 397: puffer, pufferfish, blowfish, globefish\par}
\end{minipage}

\vspace{0.15em}

\begin{minipage}[t]{0.495\textwidth}
\centering
\includegraphics[width=\linewidth]{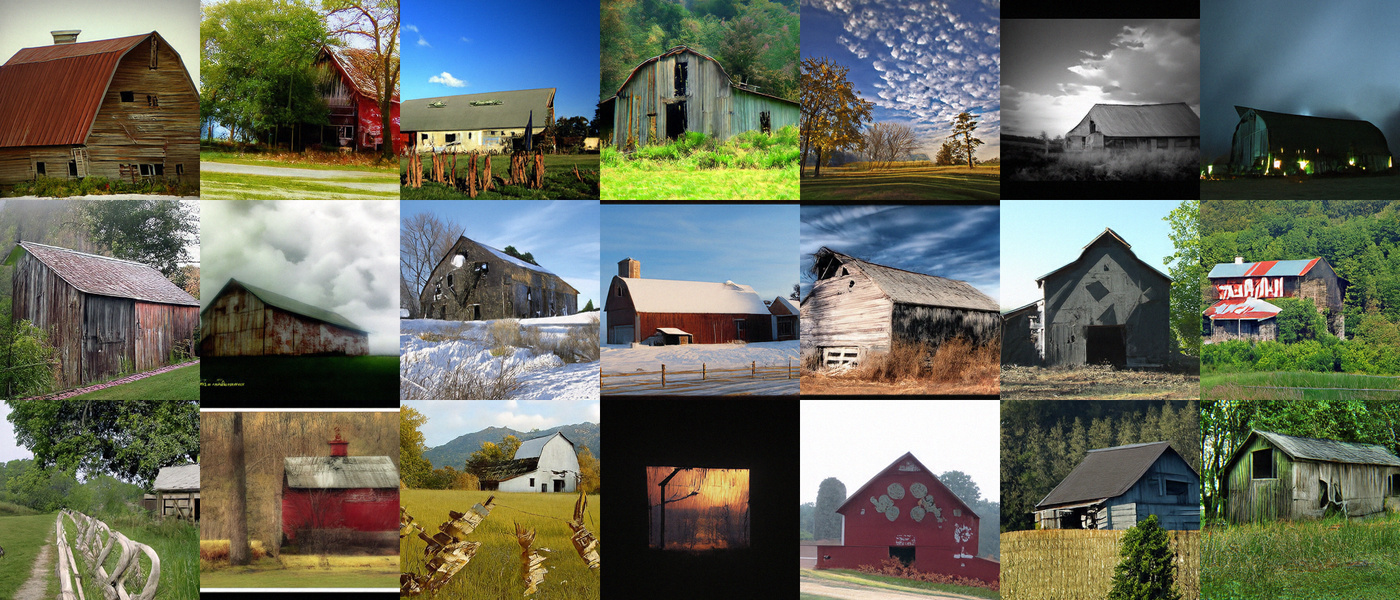}
\par\vspace{-0.25em}
{\scriptsize class 425: barn\par}
\end{minipage}\hfill
\begin{minipage}[t]{0.495\textwidth}
\centering
\includegraphics[width=\linewidth]{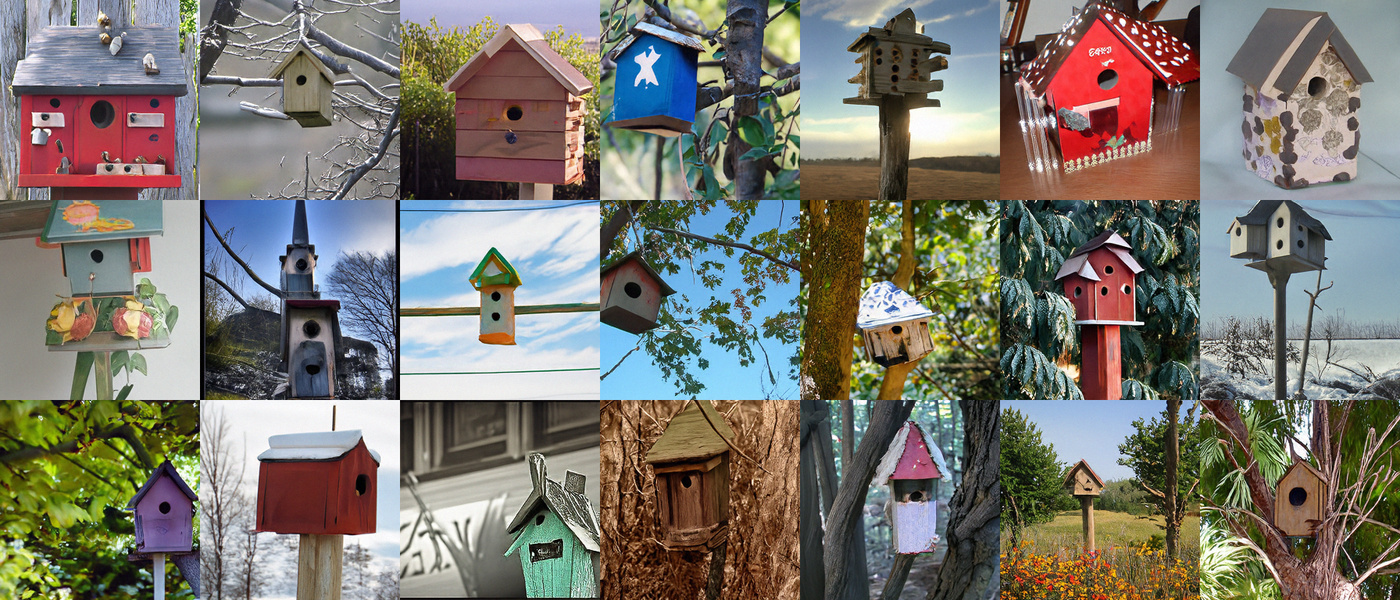}
\par\vspace{-0.25em}
{\scriptsize class 448: birdhouse\par}
\end{minipage}

\vspace{0.15em}

\begin{minipage}[t]{0.495\textwidth}
\centering
\includegraphics[width=\linewidth]{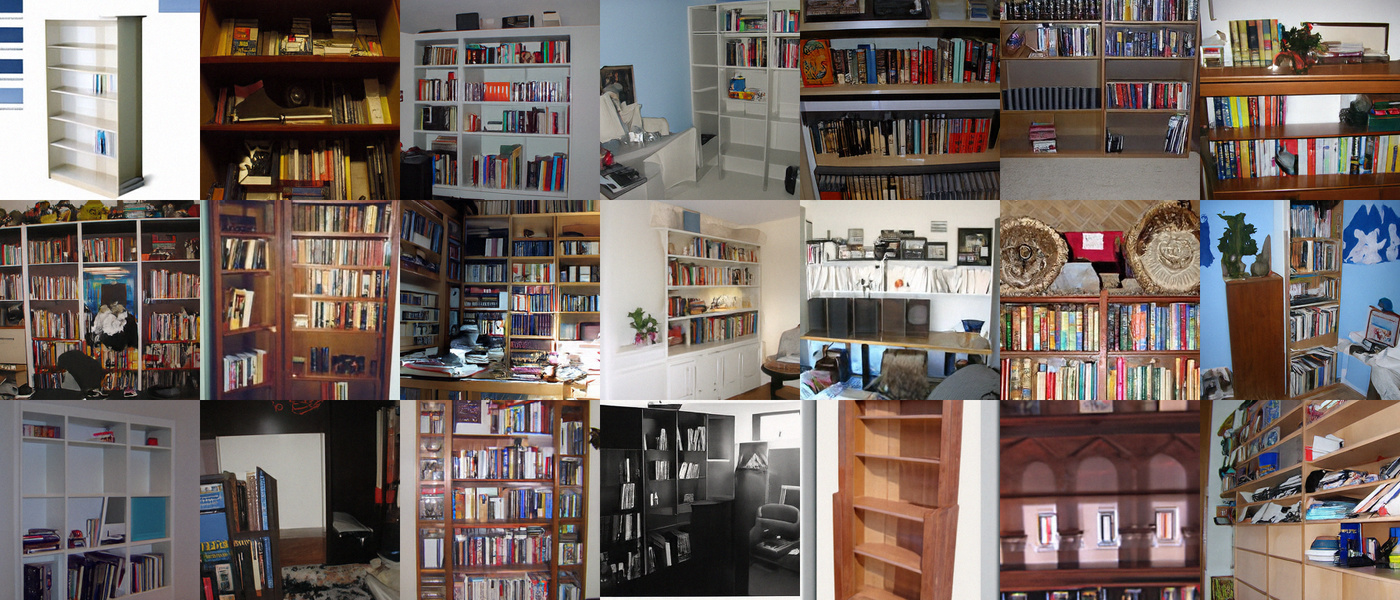}
\par\vspace{-0.25em}
{\scriptsize class 453: bookcase\par}
\end{minipage}\hfill
\begin{minipage}[t]{0.495\textwidth}
\centering
\includegraphics[width=\linewidth]{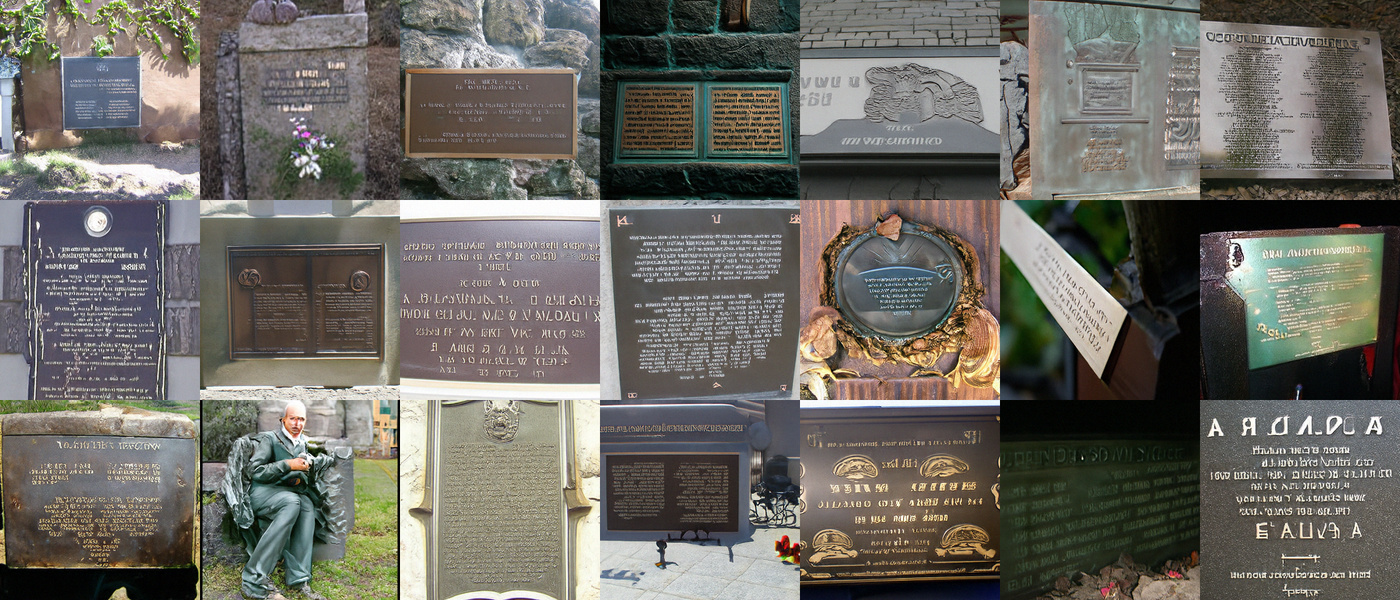}
\par\vspace{-0.25em}
{\scriptsize class 458: brass, memorial tablet, plaque\par}
\end{minipage}

\vspace{0.15em}

\begin{minipage}[t]{0.495\textwidth}
\centering
\includegraphics[width=\linewidth]{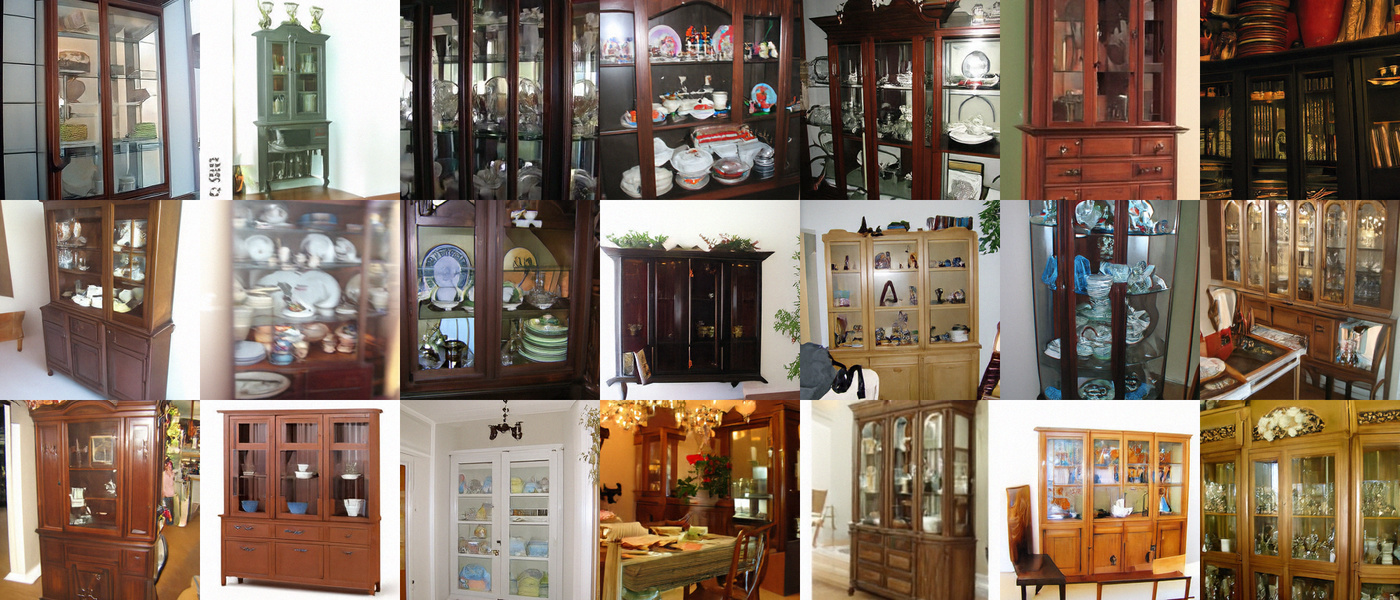}
\par\vspace{-0.25em}
{\scriptsize class 495: china cabinet, china closet\par}
\end{minipage}\hfill
\begin{minipage}[t]{0.495\textwidth}
\centering
\includegraphics[width=\linewidth]{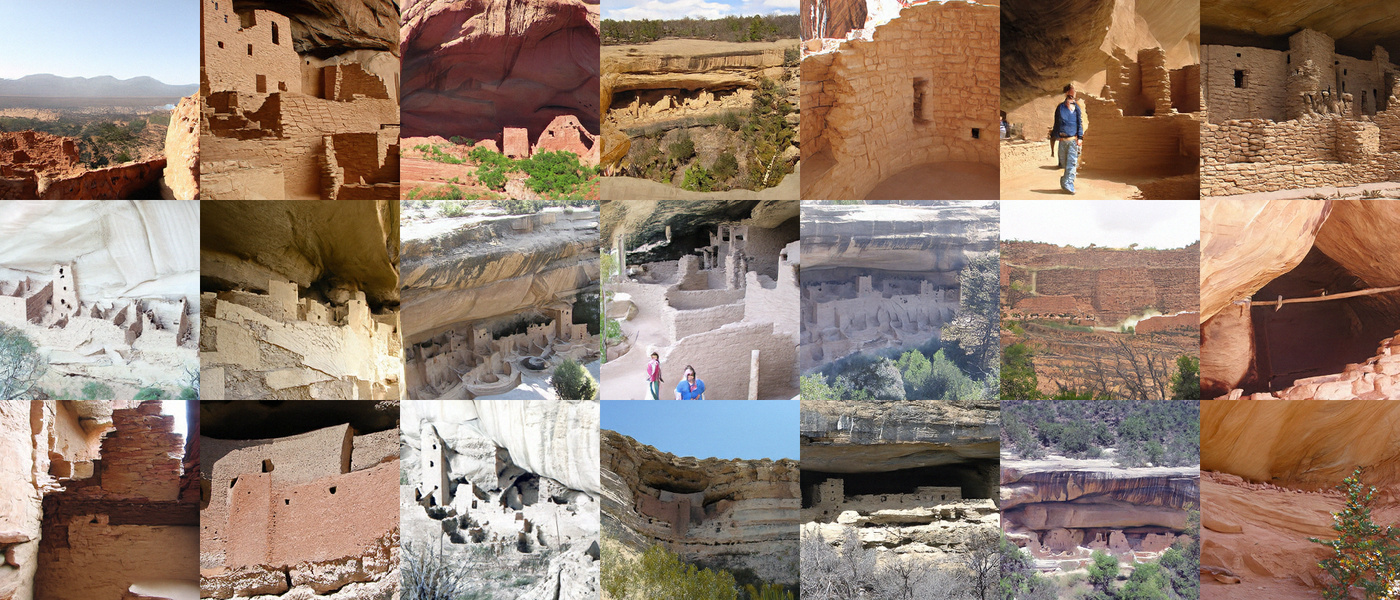}
\par\vspace{-0.25em}
{\scriptsize class 500: cliff dwelling\par}
\end{minipage}

\caption{{Uncurated class-conditional samples on ImageNet $256\times256$ using JiT-H/16 with SSG.}
The CFG and SSG scales are set to 1.9 and 1.2.}
\label{fig:ssg_h16_samples_1}
\end{figure*}

\begin{figure*}[p]
  \centering

  \begin{minipage}[t]{0.495\textwidth}
  \centering
  \includegraphics[width=\linewidth]{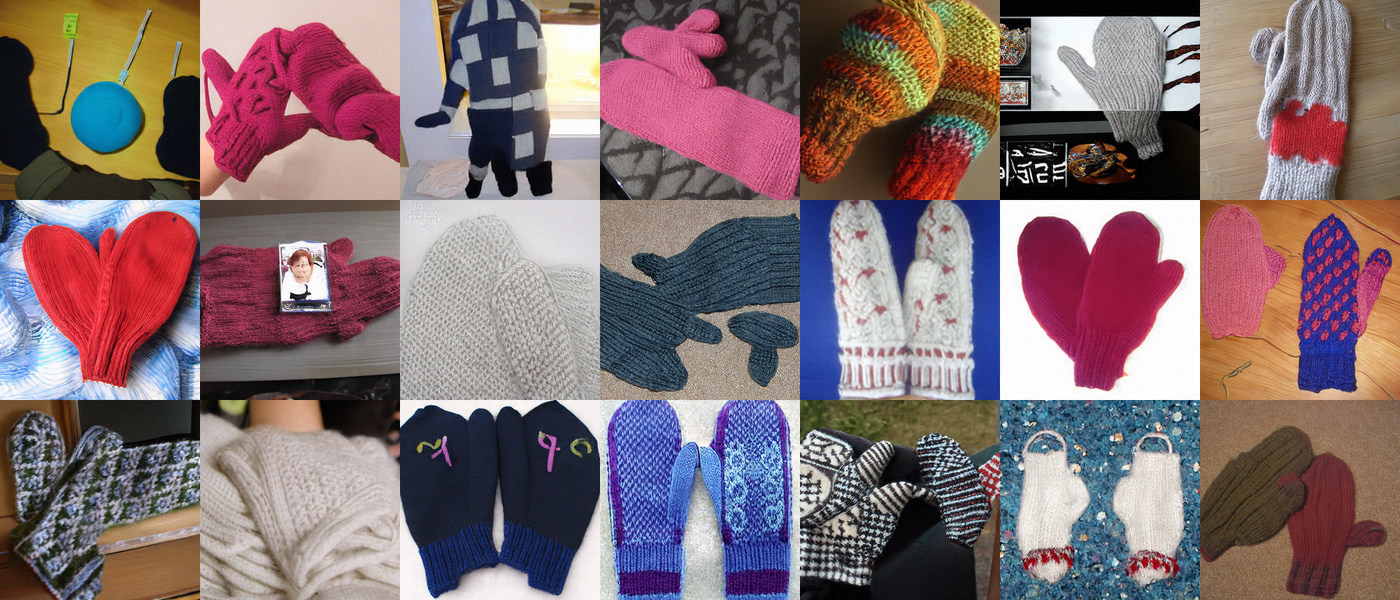}
  \par\vspace{-0.25em}
  {\scriptsize class 658: mitten\par}
  \end{minipage}\hfill
  \begin{minipage}[t]{0.495\textwidth}
  \centering
  \includegraphics[width=\linewidth]{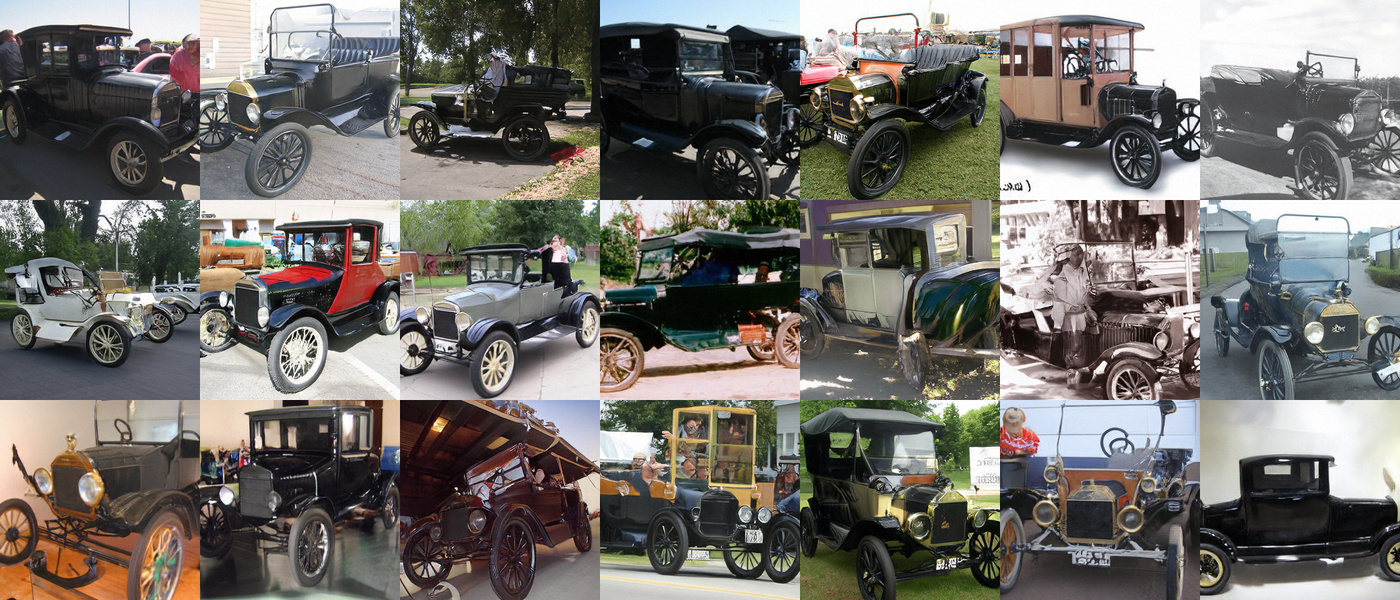}
  \par\vspace{-0.25em}
  {\scriptsize class 661: Model T\par}
  \end{minipage}

  \vspace{0.15em}

  \begin{minipage}[t]{0.495\textwidth}
  \centering
  \includegraphics[width=\linewidth]{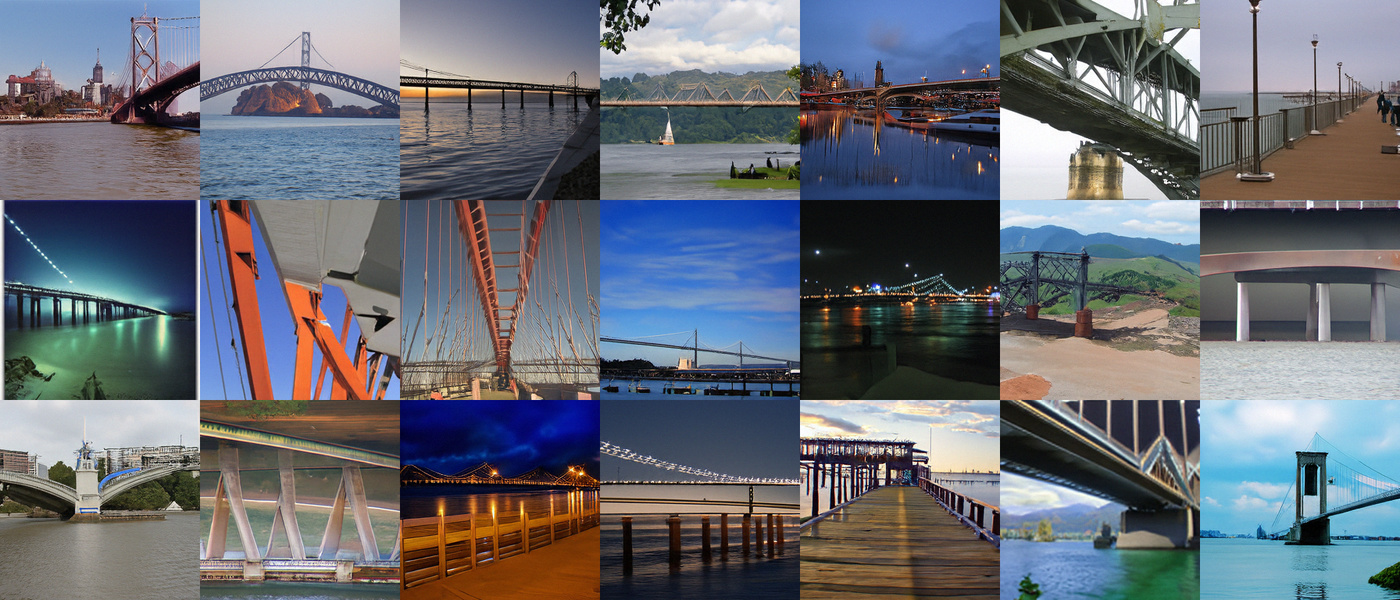}
  \par\vspace{-0.25em}
  {\scriptsize class 718: pier\par}
  \end{minipage}\hfill
  \begin{minipage}[t]{0.495\textwidth}
  \centering
  \includegraphics[width=\linewidth]{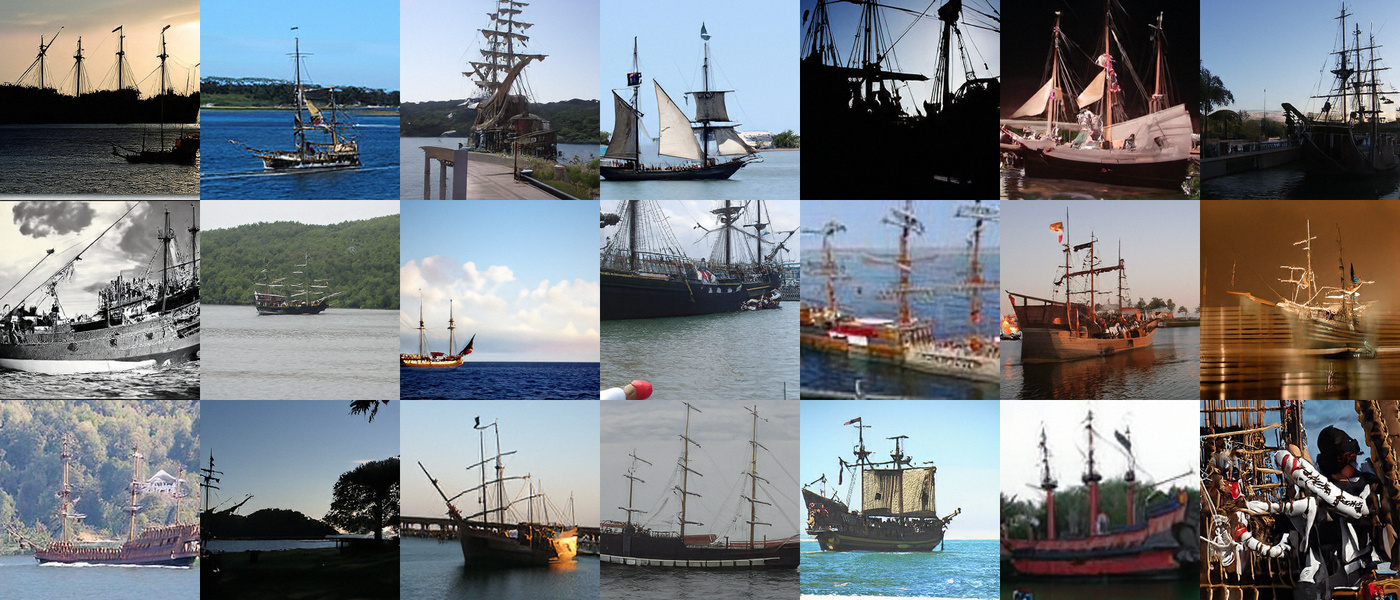}
  \par\vspace{-0.25em}
  {\scriptsize class 724: pirate, pirate ship\par}
  \end{minipage}

  \vspace{0.15em}

  \begin{minipage}[t]{0.495\textwidth}
  \centering
  \includegraphics[width=\linewidth]{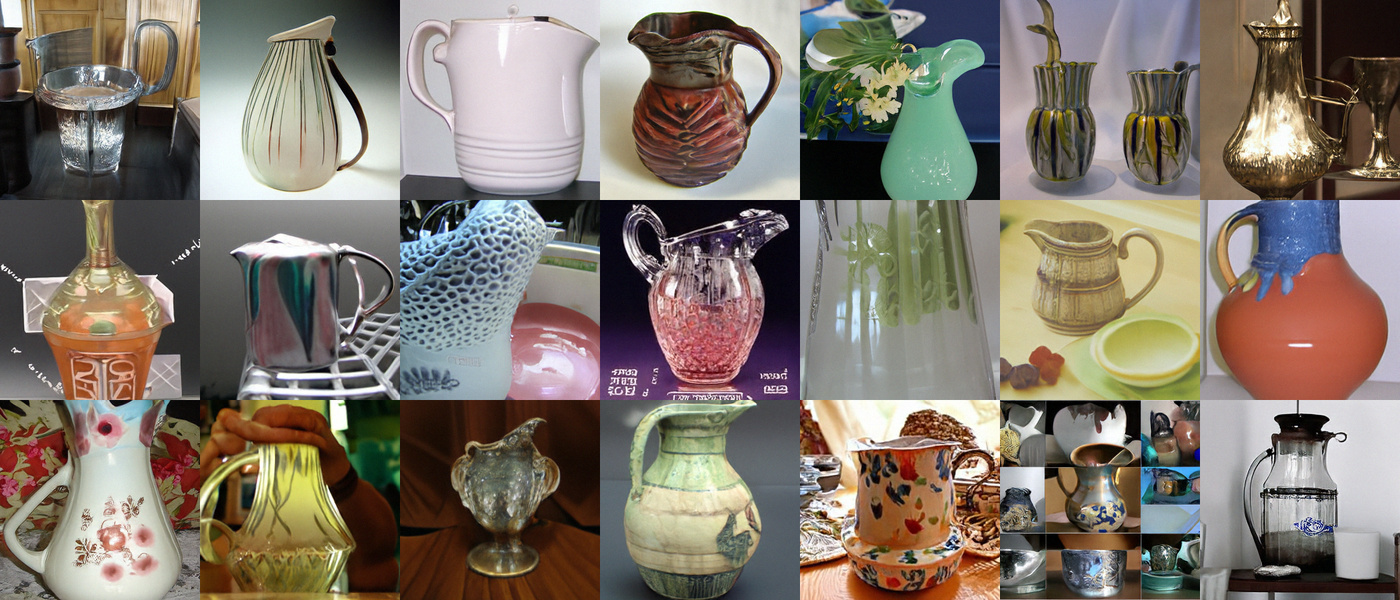}
  \par\vspace{-0.25em}
  {\scriptsize class 725: pitcher, ewer\par}
  \end{minipage}\hfill
  \begin{minipage}[t]{0.495\textwidth}
  \centering
  \includegraphics[width=\linewidth]{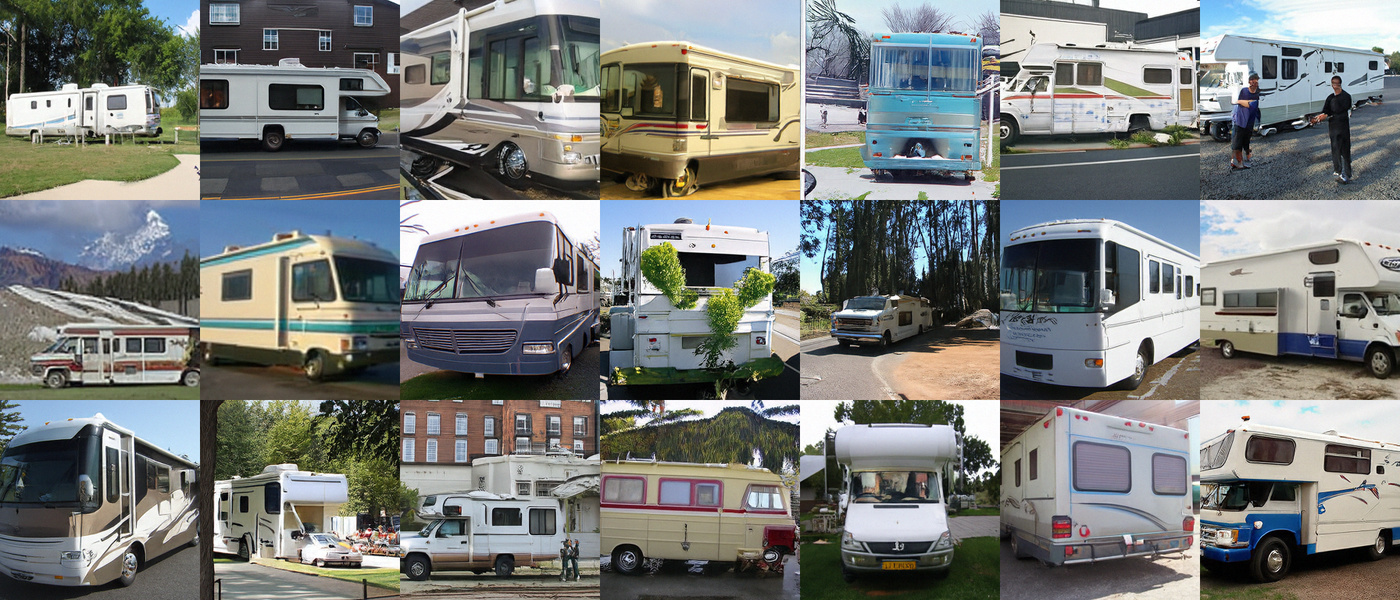}
  \par\vspace{-0.25em}
  {\scriptsize class 757: recreational vehicle, RV, R.V.\par}
  \end{minipage}

  \vspace{0.15em}

  \begin{minipage}[t]{0.495\textwidth}
  \centering
  \includegraphics[width=\linewidth]{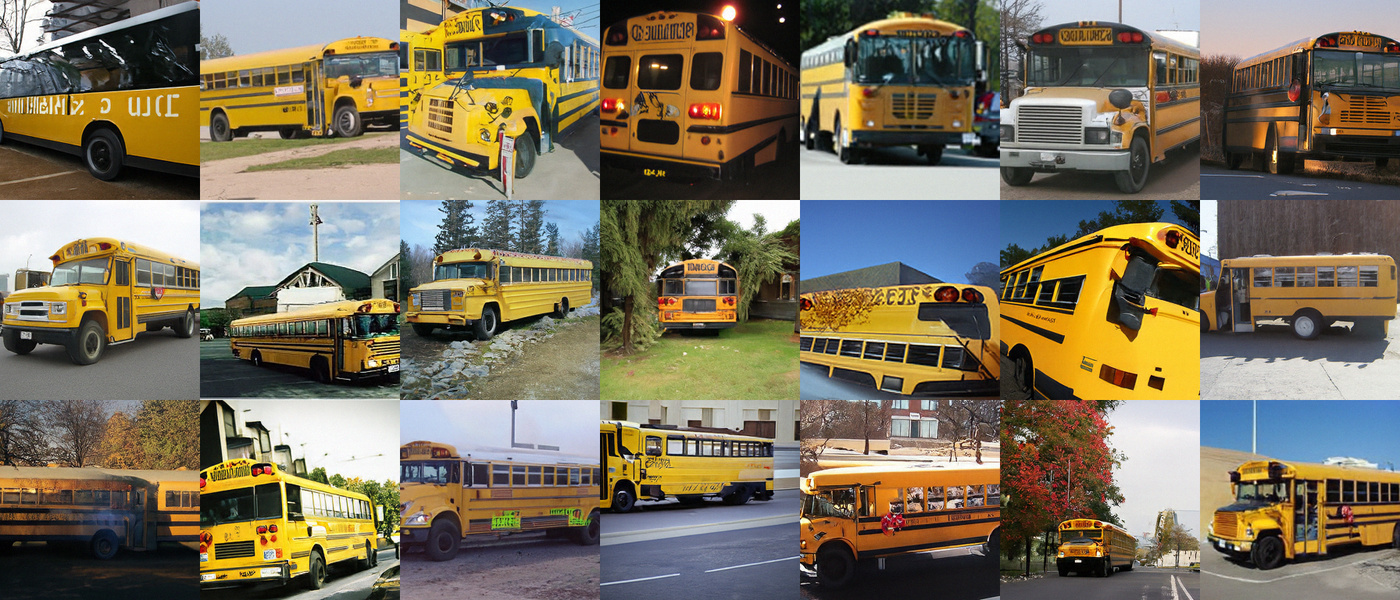}
  \par\vspace{-0.25em}
  {\scriptsize class 779: school bus\par}
  \end{minipage}\hfill
  \begin{minipage}[t]{0.495\textwidth}
  \centering
  \includegraphics[width=\linewidth]{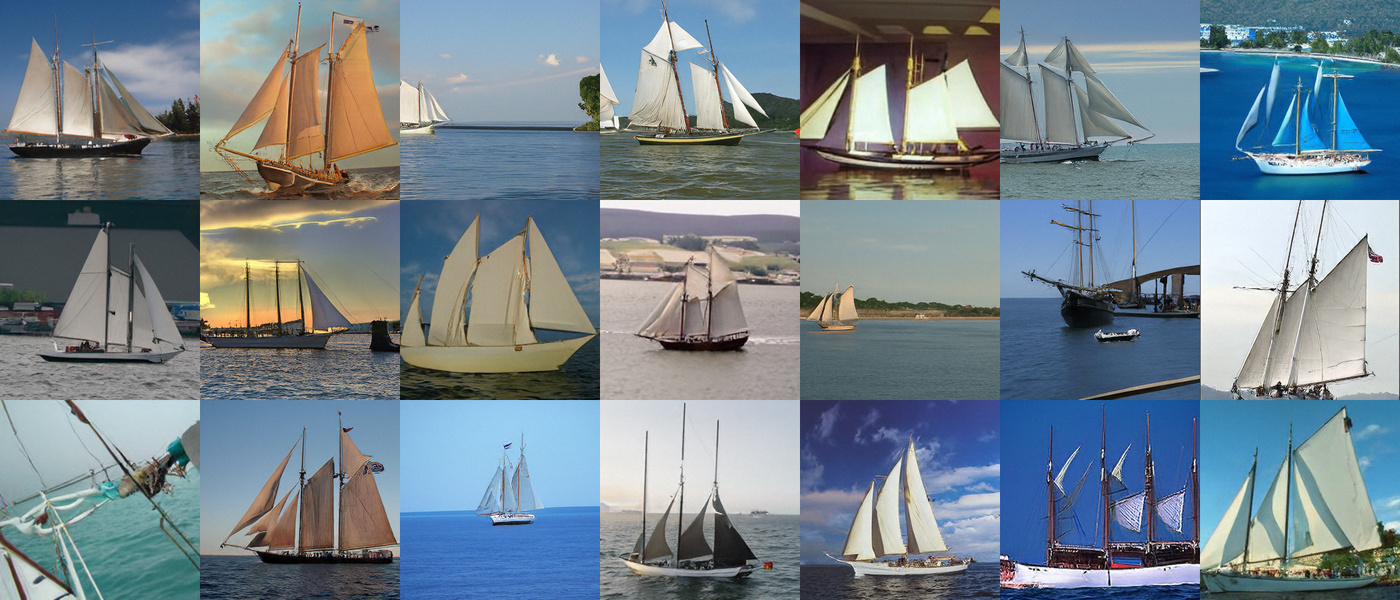}
  \par\vspace{-0.25em}
  {\scriptsize class 780: schooner\par}
  \end{minipage}

  \vspace{0.15em}

  \begin{minipage}[t]{0.495\textwidth}
  \centering
  \includegraphics[width=\linewidth]{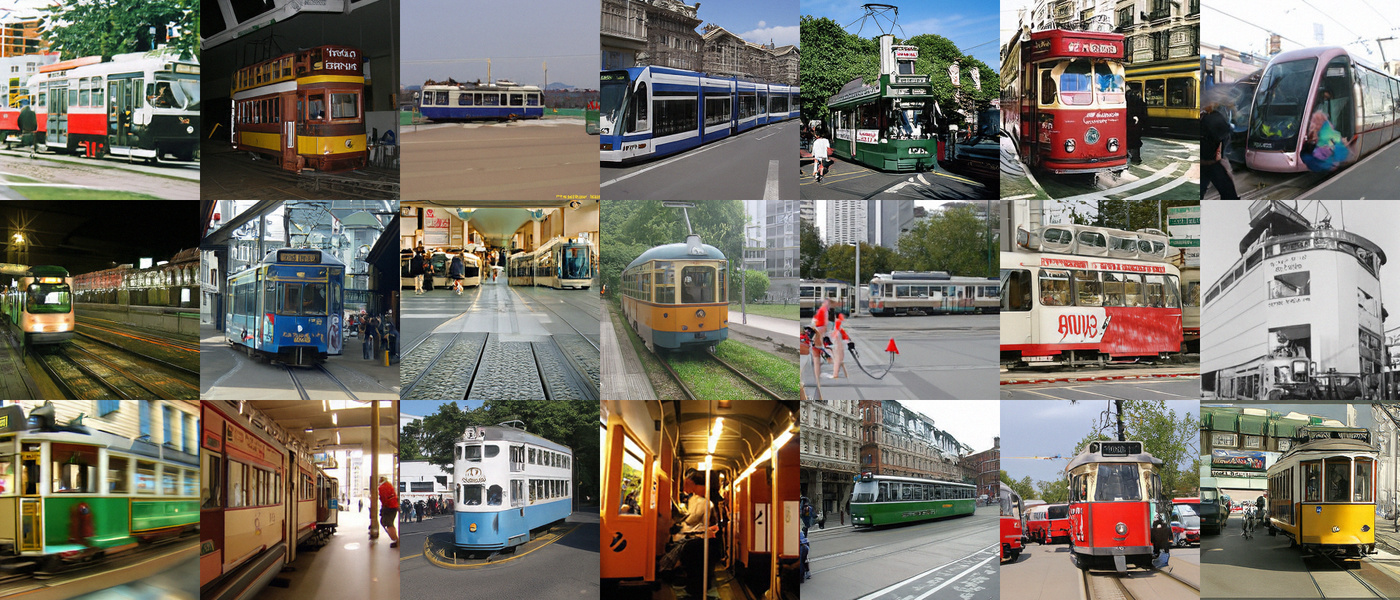}
  \par\vspace{-0.25em}
  {\scriptsize class 829: streetcar, tram, tramcar, trolley, trolley car\par}
  \end{minipage}\hfill
  \begin{minipage}[t]{0.495\textwidth}
  \centering
  \includegraphics[width=\linewidth]{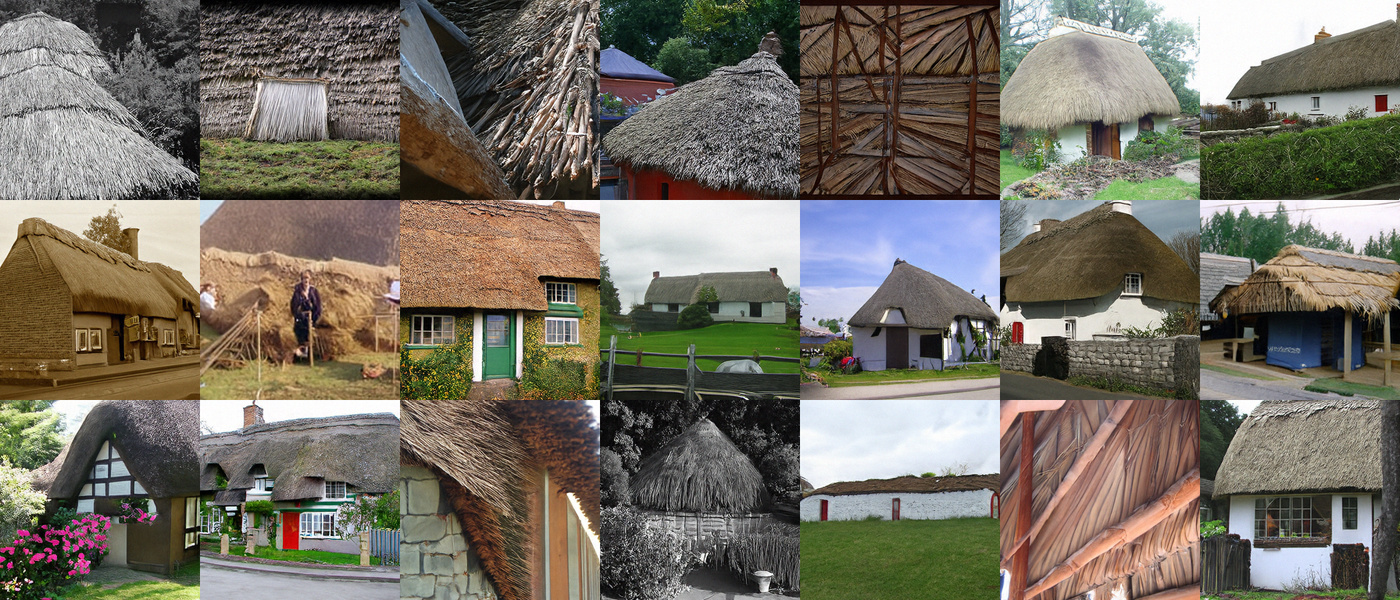}
  \par\vspace{-0.25em}
  {\scriptsize class 853: thatch, thatched roof\par}
  \end{minipage}

  \caption{{Uncurated class-conditional samples on ImageNet $256\times256$ using JiT-H/16 with SSG.}
The CFG and SSG scales are set to 1.9 and 1.2.}
  \label{fig:ssg_h16_samples_2}
  \end{figure*}

\begin{figure*}[p]
  \centering

  \begin{minipage}[t]{0.495\textwidth}
  \centering
  \includegraphics[width=\linewidth]{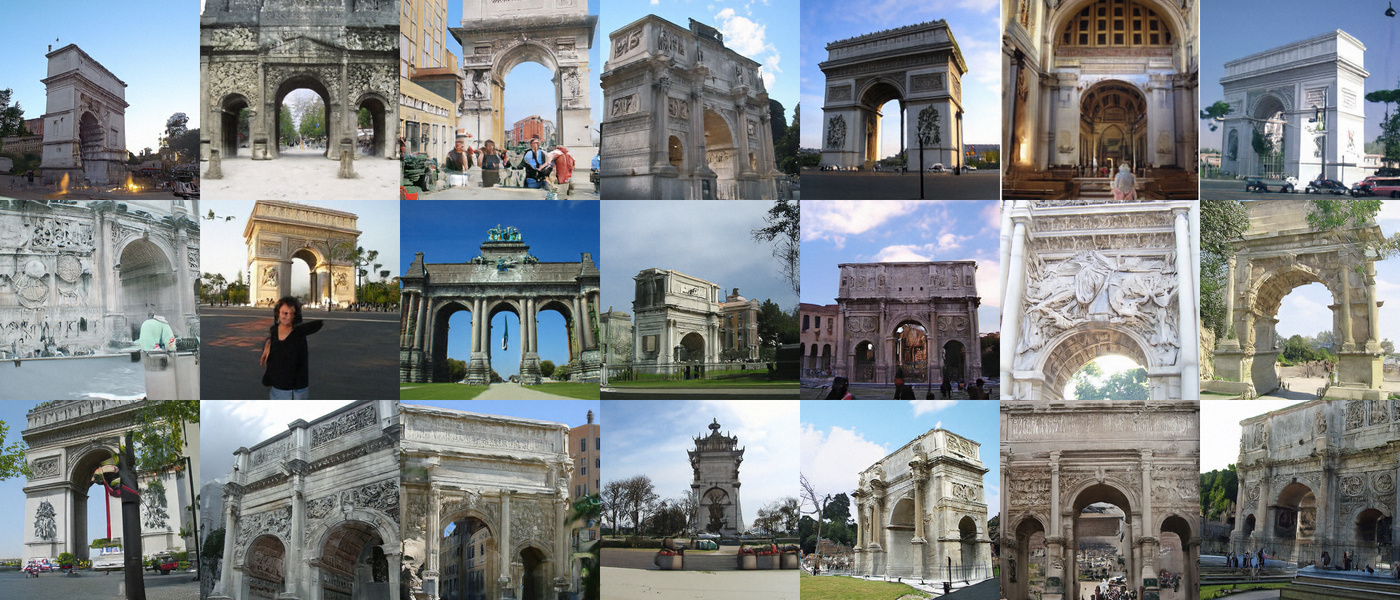}
  \par\vspace{-0.25em}
  {\scriptsize class 873: triumphal arch\par}
  \end{minipage}\hfill
  \begin{minipage}[t]{0.495\textwidth}
  \centering
  \includegraphics[width=\linewidth]{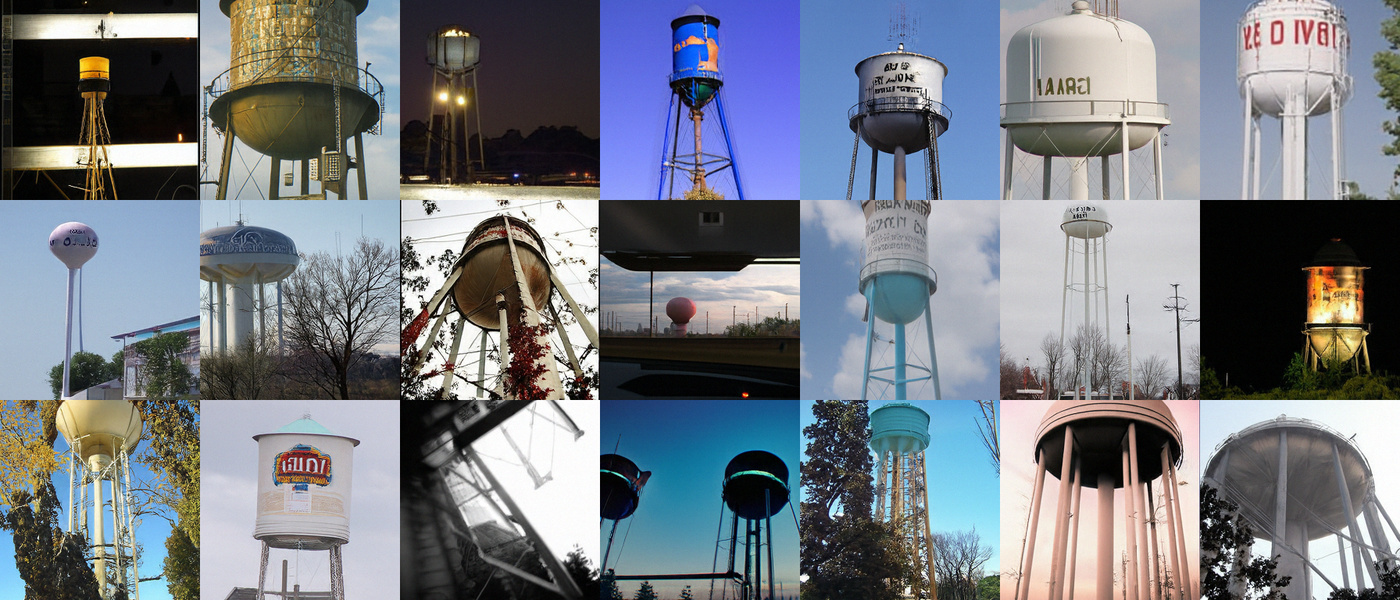}
  \par\vspace{-0.25em}
  {\scriptsize class 900: water tower\par}
  \end{minipage}

  \vspace{0.15em}

  \begin{minipage}[t]{0.495\textwidth}
  \centering
  \includegraphics[width=\linewidth]{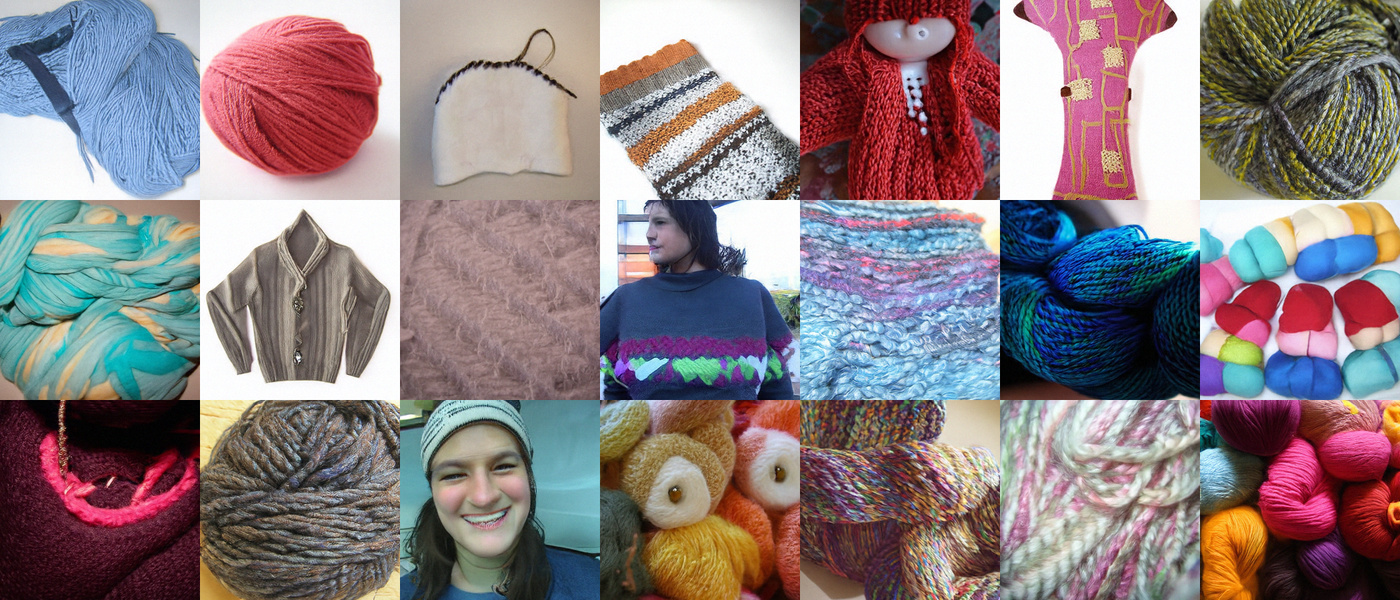}
  \par\vspace{-0.25em}
  {\scriptsize class 911: wool, woolen, woollen\par}
  \end{minipage}\hfill
  \begin{minipage}[t]{0.495\textwidth}
  \centering
  \includegraphics[width=\linewidth]{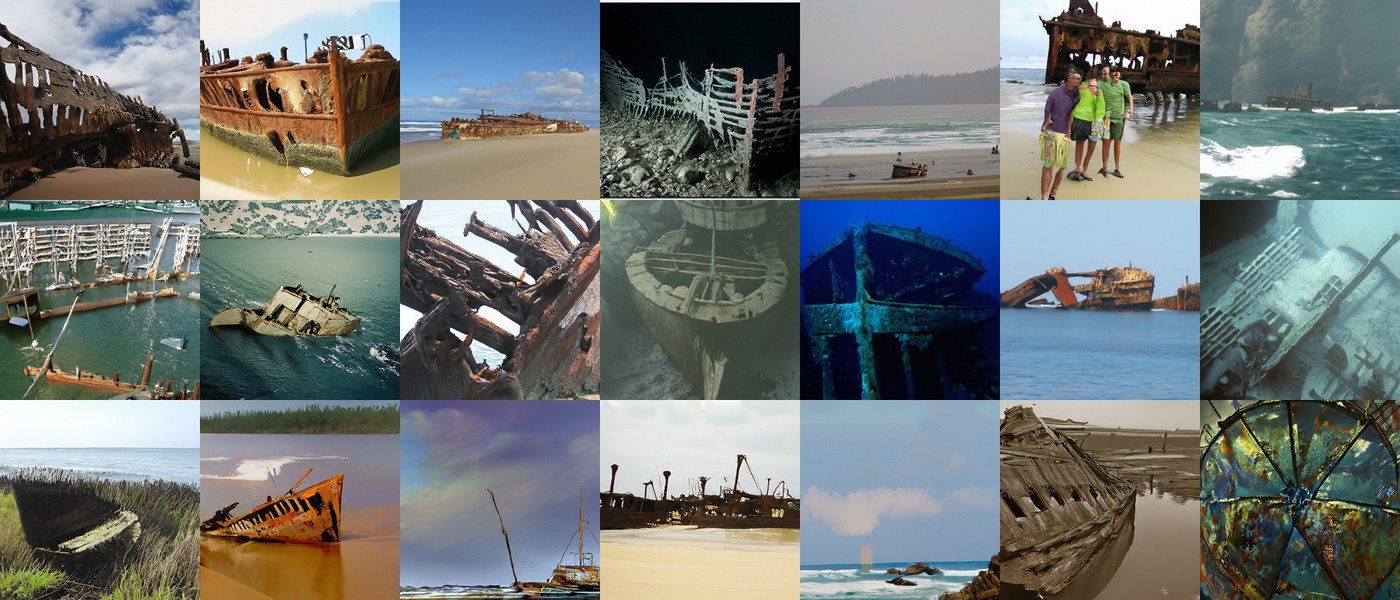}
  \par\vspace{-0.25em}
  {\scriptsize class 913: wreck\par}
  \end{minipage}

  \vspace{0.15em}

  \begin{minipage}[t]{0.495\textwidth}
  \centering
  \includegraphics[width=\linewidth]{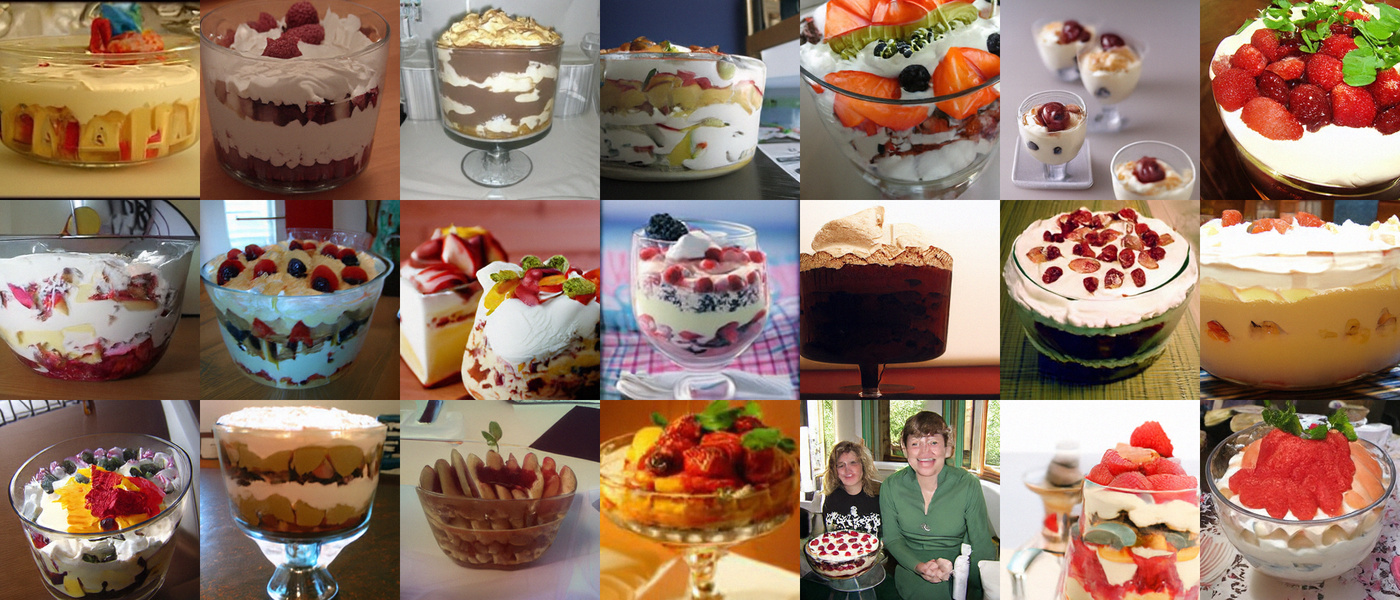}
  \par\vspace{-0.25em}
  {\scriptsize class 927: trifle\par}
  \end{minipage}\hfill
  \begin{minipage}[t]{0.495\textwidth}
  \centering
  \includegraphics[width=\linewidth]{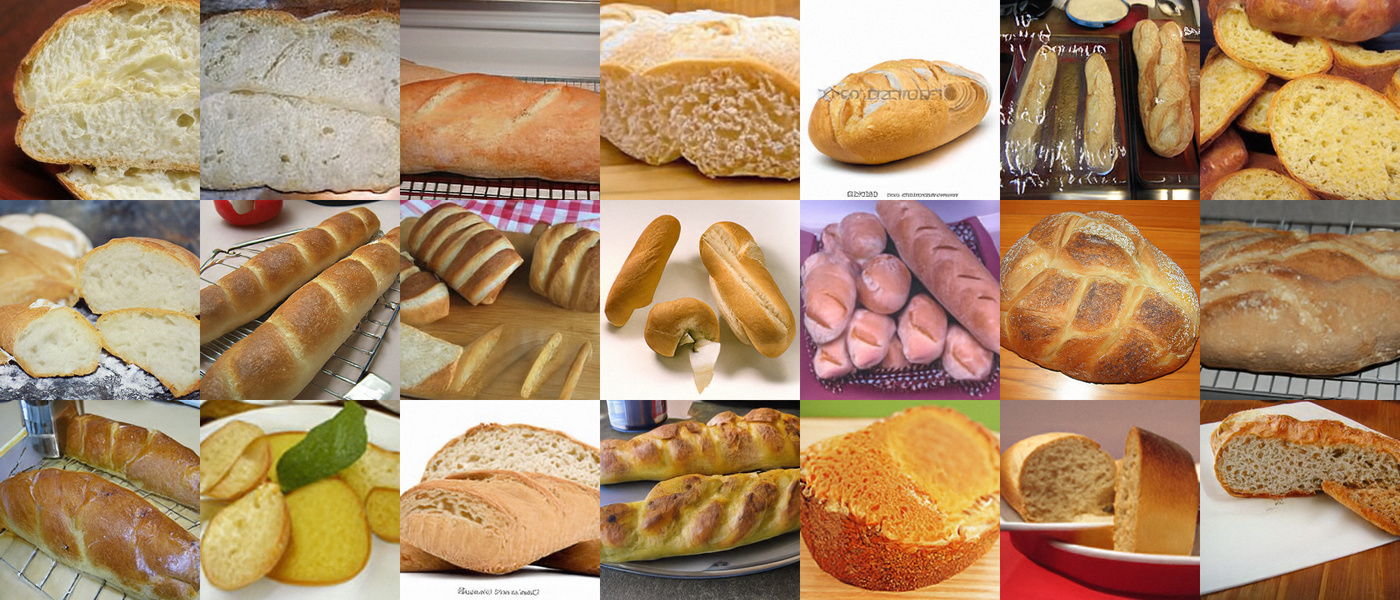}
  \par\vspace{-0.25em}
  {\scriptsize class 930: French loaf\par}
  \end{minipage}

  \vspace{0.15em}

  \begin{minipage}[t]{0.495\textwidth}
  \centering
  \includegraphics[width=\linewidth]{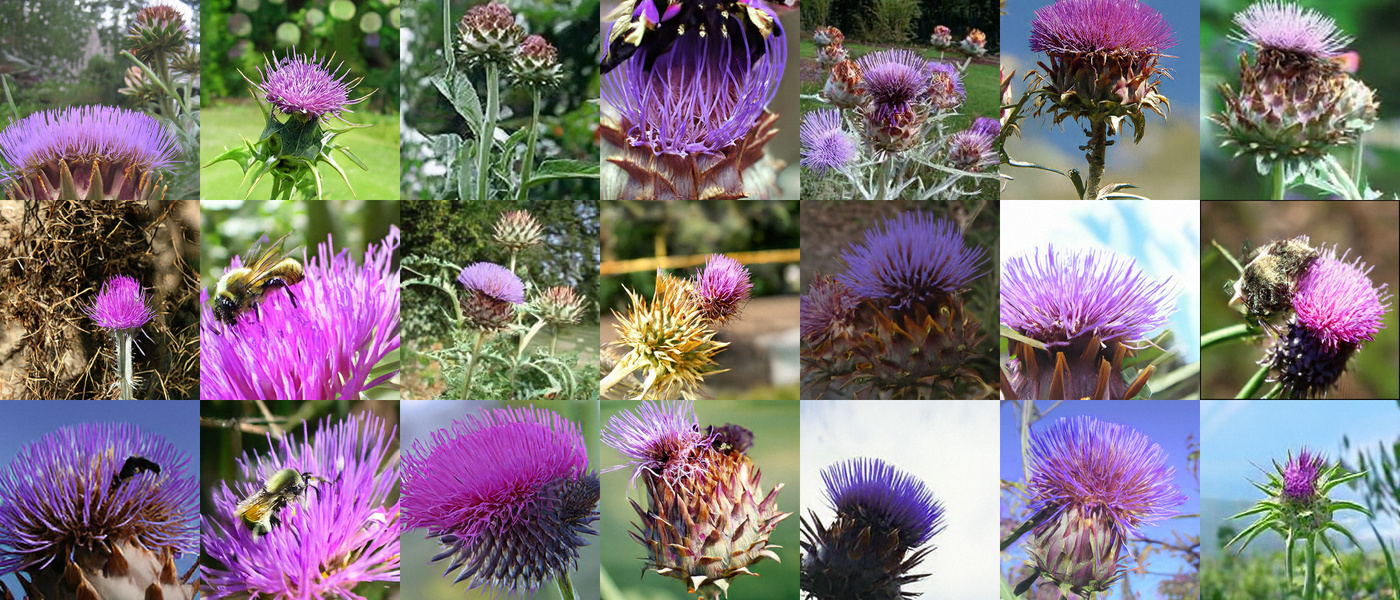}
  \par\vspace{-0.25em}
  {\scriptsize class 946: cardoon\par}
  \end{minipage}\hfill
  \begin{minipage}[t]{0.495\textwidth}
  \centering
  \includegraphics[width=\linewidth]{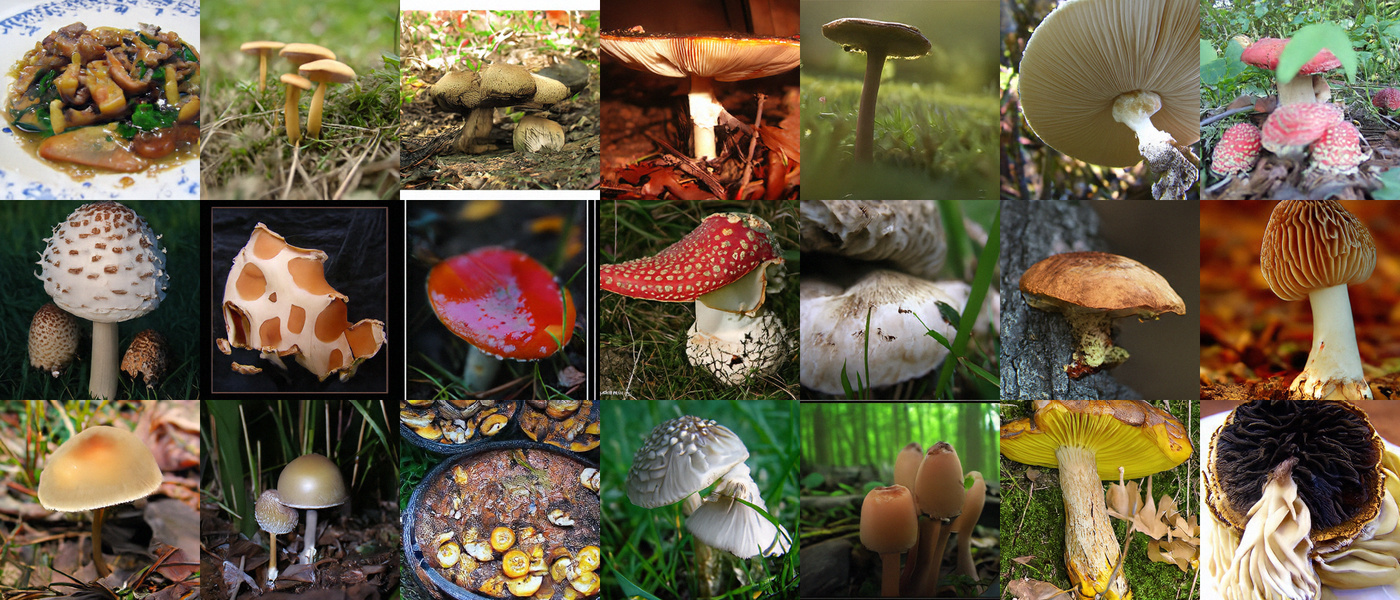}
  \par\vspace{-0.25em}
  {\scriptsize class 947: mushroom\par}
  \end{minipage}

  \vspace{0.15em}

  \begin{minipage}[t]{0.495\textwidth}
  \centering
  \includegraphics[width=\linewidth]{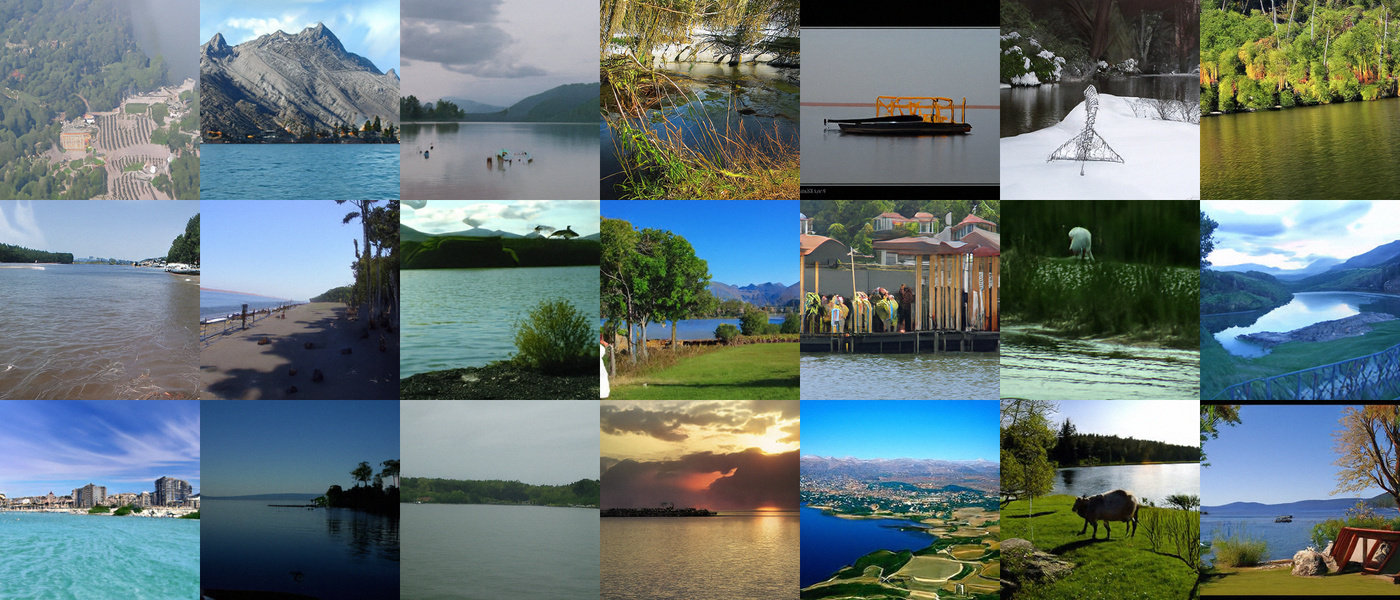}
  \par\vspace{-0.25em}
  {\scriptsize class 975: lakeside, lakeshore\par}
  \end{minipage}\hfill
  \begin{minipage}[t]{0.495\textwidth}
  \centering
  \includegraphics[width=\linewidth]{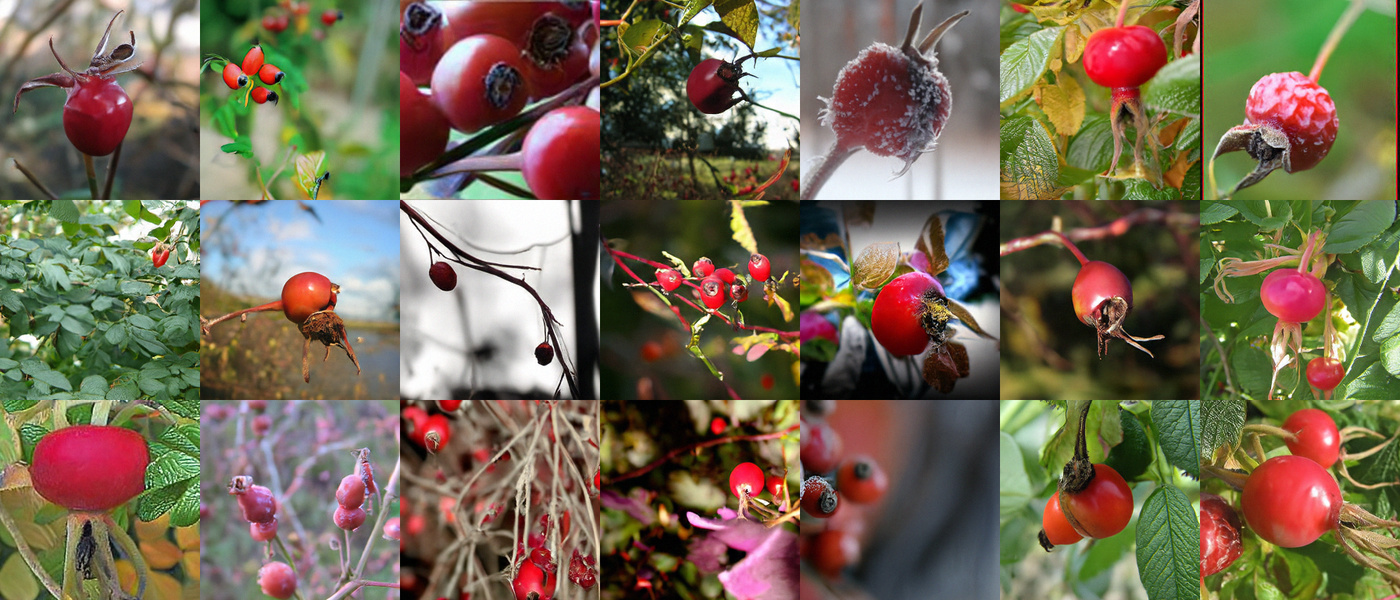}
  \par\vspace{-0.25em}
  {\scriptsize class 989: hip, rose hip, rosehip\par}
  \end{minipage}

  \caption{{Uncurated class-conditional samples on ImageNet $256\times256$ using JiT-H/16 with SSG.}
The CFG and SSG scales are set to 1.9 and 1.2.}
  \label{fig:ssg_h16_samples_3}
  \end{figure*}

We provide additional class-conditional samples generated by JiT-H/16 with SSG. As seen in Fig. \ref{fig:ssg_h16_samples_1}, Fig. \ref{fig:ssg_h16_samples_2}, and Fig. \ref{fig:ssg_h16_samples_3}, our method generates high-quality samples. 
 \begin{table*}[t!]
  \centering
  {\small
  \setlength{\tabcolsep}{6pt}
  \renewcommand{\arraystretch}{1.15}
  \begin{tabular}{@{}lccccc@{}}
  \toprule
  & JiT-B & JiT-L & JiT-H & PixelREPA-H & DeCo-XL \\
  \midrule

  \multicolumn{6}{c}{\textbf{Architecture}} \\
  \midrule
  Backbone parameters
  & 131M & 459M & 953M & 953M & 682M \\

  Backbone depth
  & 12 & 24 & 32 & 32 & 28 (enc.) / 3 (dec.) \\

  Hidden size
  & 768 & 1024 & 1280 & 1280 & 1152 (enc.) / 32 (dec.) \\

  Image size
  & \multicolumn{3}{c}{256 / 512} & 256 & 256 \\

  Patch size
  & \multicolumn{3}{c}{16 / 32} & 16 & 16 \\

  \midrule
  \multicolumn{6}{c}{\textbf{Adapter}} \\
  \midrule
  Attachment layer
  & 6 & 6 & 8 & 8 & 8 (enc.) \\

  Adapter depth
  & 1 block & 1 block & 2 blocks & 2 blocks & 2 enc.\ blocks \\

  % Initialization
  % & Block 12 & Block 24 & Blocks 31--32 & Blocks 31--32 & Blocks 27--28 \\

  Output layer
  & \multicolumn{4}{c}{Linear layer} & Pixel decoder \\

  Prediction target
  & \multicolumn{4}{c}{Clean image} & Velocity \\

  Adapter parameters
  & 12M & 22M & 63M & 63M & 57M \\

  \midrule
  \multicolumn{6}{c}{\textbf{Adapter Training}} \\
  \midrule
  Epochs
  & 30 & 30 & 50 & 50 & 50 \\

  Synthetic samples
  & \multicolumn{5}{c}{1M} \\

  Optimizer
  & \multicolumn{5}{c}{Adam, $\beta_1=0.9$, $\beta_2=0.95$} \\

  Global batch size
  & \multicolumn{4}{c}{1024} & 256 \\

  Learning rate
  & \multicolumn{4}{c}{2e-4} & 5e-5 \\

  % Warmup epochs
  % & \multicolumn{5}{c}{0} \\

  % Adapter EMA
  % & \multicolumn{5}{c}{0.9993} \\

  % Timestep distribution
  % & \multicolumn{4}{c}{$\operatorname{sigmoid}(\mathcal{N}(-0.8,0.8^2))$}
  % & $\operatorname{sigmoid}(\mathcal{N}(0,1))$ \\

  % Condition dropout
  % & \multicolumn{4}{c}{0.1} & 0.2 \\

  \midrule
  \multicolumn{6}{c}{\textbf{Sampling}} \\
  \midrule
  Sampler
  & \multicolumn{5}{c}{Heun} \\

  Sampling steps
  & \multicolumn{5}{c}{50} \\

  Noise scale
  & \multicolumn{3}{c}{1.0 / 2.0} & 1.0 & 1.0 \\

  SSG interval w/o CFG
  & \multicolumn{5}{c}{$[0,1]$} \\

  CFG and SSG interval
  & \multicolumn{5}{c}{$[0.1,1.0]$} \\

  \bottomrule
  \end{tabular}
  }
  \caption{
  Backbone, adapter, training, and sampling configurations.
  Backbone parameters exclude the adapter, whose parameter count includes its output layer.
  For JiT, the 256 and 512 resolutions use patch sizes 16 and 32 and noise scales 1.0 and 2.0, respectively.
  All backbones remain frozen during adapter training.
  }
  \label{tab_supp_implementation}
  \end{table*}
  
\section{Limitations and Future Work}

Although SSG is effective, our experiments mainly focus on class-conditional ImageNet generation with pixel-space diffusion models.
It remains unclear whether the intermediate-to-final refinement observed in these models also generalizes to larger text-conditioned models or domains with different image distributions.
Future work could extend SSG to text-to-image generation, broader datasets, and other diffusion architectures.

\bibliography{aaai2027}


\end{document}